\documentclass[sigconf]{aamas} 

\usepackage{balance} 
\usepackage[toc,page]{appendix}
\usepackage{xcolor}
\renewcommand\footnotetextcopyrightpermission[1]{} 

\setcopyright{none}

\title[AAMAS-2021 Formatting Instructions]{Loss Bounds for Approximate Influence-Based Abstraction}

\author{Elena Congeduti}
\affiliation{
	\department{Computer Science}
  \institution{Delft University of Technology}}
\email{E.Congeduti@tudelft.nl}

\author{Alexander Mey}
\affiliation{
	\department{Computer Science}
	\institution{Delft University of Technology}}
\email{A.Mey@tudelft.nl}

\author{Frans A. Oliehoek}
\affiliation{
	\department{Computer Science}
	\institution{Delft University of Technology}}
\email{F.A.Oliehoek@tudelft.nl}

\begin{abstract}
Sequential decision making techniques hold great promise to improve the performance of many real-world systems, but computational
complexity hampers their principled application.
Influence-based abstraction aims to gain leverage by modeling local subproblems together with the `influence' that the rest of the system exerts on them.
While computing exact representations of such influence might be intractable, learning approximate representations offers a promising approach to enable scalable solutions. This paper investigates the performance of such approaches from a theoretical perspective. The
primary contribution is the derivation of sufficient conditions on approximate influence representations that can guarantee solutions with
small value loss. In particular we show that neural networks trained with cross entropy are well suited to learn approximate influence representations.
Moreover, we provide a sample based formulation of the bounds, which reduces the gap to applications. Finally, driven by
our theoretical insights, we propose approximation error estimators, which empirically reveal to correlate well with the value loss.
\end{abstract}


\newcommand{\BibTeX}{\rm B\kern-.05em{\sc i\kern-.025em b}\kern-.08em\TeX}

\usepackage{amsmath}
\usepackage{wrapfig}
\usepackage{graphicx}
\usepackage{algorithm}
\usepackage{algorithmic}
\usepackage{mathtools}
\usepackage{enumitem}
\usepackage{comment}
\usepackage{bbm}
\usepackage{xcolor}
\usepackage{amsthm}
\usepackage{subcaption}

\usepackage{cleveref}

\providecommand{\defas}{\mathrel{\triangleq}}

\theoremstyle{definition}
\newtheorem{definition}{Definition}
\newtheorem{theorem}{Theorem}
\newtheorem{corollary}{Corollary}
\newtheorem{lemma}{Lemma}
\newtheorem{observation}{Observation}

\newtheorem{innercustomthm}{Theorem}
\newenvironment{customthm}[1]
{\renewcommand\theinnercustomthm{#1}\innercustomthm}
{\endinnercustomthm}

\newtheorem{innercustomlemma}{Lemma}
\newenvironment{customlemma}[1]
{\renewcommand\theinnercustomlemma{#1}\innercustomlemma}
{\endinnercustomlemma}

\newtheorem{innercustomcor}{Corollary}
\newenvironment{customcor}[1]
{\renewcommand\theinnercustomcor{#1}\innercustomcor}
{\endinnercustomcor}

\begin{document}


\pagestyle{plain}
\pagenumbering{arabic}
\fancyhead{}


\maketitle 


\section{Introduction}
Sequential decision making methods have the potential to improve control in distributed or networked systems that can involve many agents and complex environments. However, applying these methods in a principled manner is often difficult due to computational complexity. One idea to improve scalability is using abstractions, compressed representations of the sufficient information for an agent to perform optimal decisions. 
Many abstraction approaches have been suggested
\cite{givan2000bounded,li2006towards,bai2016markovian} but in order to ensure small value loss, they only allow abstracting states of the environment with similar dynamics.
A body of work on localized abstractions \cite{witwicki2010influence, becker2004decentralized,becker2004solving,varakantham2009exploiting} tries to overcome this issue, allowing to abstract large part of the system away. These approaches leverage the sparse interaction structures of many real-world domains where the agent's reward and observations depend directly only on few (local) state variables.

\textit{Influence-based abstraction} (IBA)  \cite{oliehoek2019sufficient} generalizes over these approaches and provides a unified framework for localized abstractions for a general class of multiagent problems. The basic idea is to decompose structured multiagent systems into small submodels for each agent where only few local variables are included. The `influence' summarizes the indirect effects of other agents policies and the rest of the environment on the local variables of a single agent. 
Unlike existing methods to compute best responses 
\cite{Nair03IJCAI,oliehoek2014best,Panella17JAAMAS} which involve reasoning
over the entire state space and other agents action-observation histories, IBA ensures that
the decision maker needs to reason only over few local state variables. As a result, the best-response problem can be solved much more efficiently in the local model without any loss in value. This approach may not only enable speedups in best-response problems but may serve also as the foundation for scalable multiagent approaches based, for instance, on searching in the space of influences \cite{witwicki2010influence}. 

However, even though IBA provides a lossless abstraction, in general, deducing an \emph{exact representation} of the influence itself is computationally intractable even for simple problems. Given the rapid progress in learning methods for sequence prediction \cite{karim2017lstm, lea2017temporal,bahdanau2016actor}, we envision that inducing \emph{approximate representations} of the influence offers a promising approach to scalable solutions. 
This raises a fundamental question: to what extent are such existing methods (and their proxy-loss functions) aligned with the actual influence-base abstraction objective of achieving near-optimal value?

In this work we address this issue, showing that there is theoretical and empirical evidence that employing existing sequence predictors can lead to good approximate influence representations. In particular, we derive a performance loss bound as well as a probabilistic version that provides quality guarantees on approximate influence to ensure near-optimal solutions. Like bounds based on the Bellman error, our bounds themselves are computationally intractable for large problems. This is to be expected: the performance loss is characterized by the quality of each possible prediction. Therefore, deriving an exact bound essentially corresponds to computing the exact generalization error of the employed machine learning model. Nevertheless, we show that a neural network trained with cross-entropy loss seems well suited for the prediction task, as the training objective is aligned with the bound we derived. Finally, we empirically demonstrate that it is possible to compute statistics, based on the empirical test error of approximate influence, that correlate well with the actual value loss.

In summary, our contributions in this paper are:

\begin{enumerate}[nolistsep,leftmargin=*]
	\item  Sufficient conditions for the quality of the influence approximation in terms of value loss.
	\item Discussing how these conditions are aligned with usual optimization of neural networks trained with cross entropy loss.
	\item An empirical evaluation showing that optimizing cross entropy loss of influence predictors leads to lower performance loss.
\end{enumerate}
\section{Related work}
\label{related work}

Abstraction has been widely studied as a technique to accelerate learning and improve scalability in complex (multiagent) problems. 
Many abstraction approaches have been suggested
\cite{dearden1997abstraction,givan2000bounded,li2006towards,petrik2014raam, abel2016near, dean2013model}
but they all share a common limitation: in order to guarantee bounded loss, the ground states in the same abstract states (which are clusters of ground states) must all share similar transition probabilities. However, when we want to abstract away entire sets of state variables
such guarantees will typically not be possible.
Influence-based abstraction differs essentially from these kind of abstractions. In fact, it enables abstracting entire factors still providing guarantees of optimality when the aggregated states have typically very different transition probabilities.


\section{Background}
\label{background}
First we outline the background on sequential decision making.
\subsection{Sequential Decision Making Framework}
We consider a general class of multiagent problems that can be modeled as partially observable stochastic games \cite{hansen2004dynamic}.
\begin{definition}[POSG]
	A \textit{Partially Observable Stochastic Game} is a tuple $M=(n,S,A,O,T,\Omega,R,h,b^0)$ where $n$ agents interact in the finite state space $S$. $A=A_1\times \dots \times A_n$ and $O=O_1\times \dots \times O_n$ are the finite spaces of joint actions and observations. $T$ models the transition probabilities as $T(s'|s,a)=\mathbb{P}(s' | s,a)$, namely the probability of resulting in state $s'$ when actions $a=(a_1,\dots,a_n)$ are chosen in state $s$. $\Omega$ is the observation distribution, $\Omega(o'|a,s')=\mathbb{P}(o'|a,s')$, the probability of receiving observations $o=(o_1,\dots,o_n)$ when performing actions $a$ results in state $s'$. $R(s,a)=(R_1(s,a),\dots,R_n(s,a))$ specifies the immediate reward function for each agent for taking actions $a$ in state $s$. Finally, $h$ defines the process horizon and $b^0$ the initial state distribution.
\end{definition}
In a POSG, each agent $i$ employs a policy $\pi_i$ that corresponds to a possibly stochastic map from the action-observation history to the action space. A joint policy $\pi=(\pi_1,\dots,\pi_n)$ is a tuple of policies for each of the agents. Given a joint policy $\pi$, we can define the corresponding expected cumulative reward function for an agent $i$: $V^{\pi}_i=\mathbb{E}\left [\sum_{t=1}^h R^t_i \,|\,\pi,b^0\right ]$. 

Different optimization problems and corresponding solution concepts can be considered according to the adversarial or cooperative nature of the agents. In the first case, the problem consists of searching for a Nash equilibrium (NE)\cite{Nisan07bookAGT}: a joint policy $\pi=(\pi_{i},\pi_{-i})$ such that each agent $i$ draws no advantage from deviating from its policy $\pi_i$ given the policies of the others $\pi_{-i}$. 
In cooperative settings modeled by decentralized partially observable Markov decision processes (Dec-POMDPs) 
\cite{Bernstein02MOR, Oliehoek16Book}, the solution consists of the optimal join policy $\pi$ maximizing the value function shared by all players.  
Many solutions methods for searching NE (e.g. fictitious play \cite{berger2007brown,brown1951iterative}, double oracle \cite{mcmahan2003planning}, parallel Nash memory \cite{oliehoek2006parallel}) and for optimal policies for Dec-POMDPs (e.g. influence search \cite{witwicki2010influence}) use best-response computations as their inner loop. A best response policy $\pi^*_i$ for agent $i$ against fixed policies $\pi_{-i}$ for the other agents maximizes the value function $\max_{\pi_i}V^{(\pi_i,\pi_{-i})}_i.$ 

However, computing best responses implies reasoning over the entire state space and other agents beliefs \cite{Nair03IJCAI}. To ease this task we want to exploit the structural properties of the environment to build local abstractions.
Frequently, in fact, the state space can be thought as composed of different state variables, or \textit{factors} \cite{boutilier1999decision}. In this case, the model is called 
\textit{factored POSG} and every state $s\in S$ can be represented as a tuple of factors instantiations $s=(x_1, \dots,x_m)$. This structure allows to decompose the system into (weakly) coupled subproblems, the \textit{local models}, for single agents including only few factors.
Thus, the idea is to define an equivalent problem to the best response in which the agent needs only to reason over the factors included in the local model. 
\subsection{Influence-Based Abstraction}
Influence-based abstraction \cite{oliehoek2019sufficient} 
formalizes the concept of the local model for an agent in a factored POSG and provides the formal framework of the local abstraction for the best-response problem. 
%

To simplify the discussion, we restrict to locally fully-observable POSGs \cite{Goldman04JAIR}. 
We use capital letters to denote state variables and small letters for variable instantiations.
\begin{definition}[Fully Observable Local Model]\label{FOLM}
	Given a factored POSG, a \textit{Fully Observable Local Model} for an agent is a subset of fully observable state variables, called \emph{modeled factors}, including all the state variables directly affecting the observations and reward. We denote the \emph{local state}, that is the collection of all the modeled factors, as $X$ and the complementary collection of \textit{non-modeled factors} as $Y$.
\end{definition}
The idea of the local model is to exclude those factors that are not necessary to compute the best response from the subset of variables over which the agent needs to reason and that can therefore be abstracted away. 
In a local model, we distinguish between modeled factors that are only affected by other factors and actions that are modeled, called \textit{only-locally affected factors}, and the \emph{non-locally affected factors} that are modeled factors affected by at least one factor or action of the external part. We denote by $X_{\text{loc}}$ the collection of the only-locally affected factors.
\begin{definition}\label{inf_src}
	We refer to the non-modeled factors or external actions that directly exert an influence on at least one of the modeled factor as the \textit{influence sources} $Y_{\text{src}}$.
	We define the \textit{influence destinations} $X_\text{dest}$ as the modeled factors that directly experience the influence of the external part of the system through the influence sources.
\end{definition}
Thus, the local state $X$ can be thought as $X=(X_{\text{loc}}, X_{\text{dest}})$.

In other words, the local model comprises only few modeled-factors. Some of them are influenced by the external part and therefore are called influence destinations. The non-modeled factors or actions that directly exert this influence on the modeled factors are accordingly the sources of the influence. See Section 3.2 in \cite{oliehoek2019sufficient} for the formal definitions of the general, not fully-observable, case. 

When abstracting away the non-modeled factors $Y$, the dynamics of
$X_{\text{dest}}$ become non-Markovian. Accordingly, the local state
transitions are not well defined. To define the local state dynamics, we need
to include as part of the abstract state space the history of relevant modeled
actions and factors to infer the influence sources $Y_{\text{src}}$. The
notion of \textit{d-separating set} (d-set) \cite{bishop2006pattern} formalizes
this concept. 
The d-set at time $t$, $D^t$ contains the histories of the modeled factors and actions necessary to predict the influence sources $Y_{\text{src}}^t$.
\begin{definition}[Influence Point]
	The \textit{exact influence point} (EIP) $I=(I^0,\dots,I^{h-1})$
	is a collection of conditional probability distributions
	$
	I^t(Y_{\textit{src}}^t\,|\,D^t) \defas \mathbb{P}(Y_{\textit{src}}^t\,|\,D^t,\pi_{-i},b^0)
	$
	of the influence sources $Y_{\textit{src}}^t$ given the possible instantiations of the d-set.
\end{definition}
We refer to Section 4.1 in \cite{oliehoek2019sufficient} for extensive definitions of influence point and d-separating set.
The local model dynamics is formalized by the \emph{influence-augmented local model}, a factored Markov decision process (MDP) where the state space consists of the local state augmented with the d-set.
\begin{definition}[IALM]
	Consider a fully observable local model for agent $i$ in a factored POSG $M=(n,S,A,O,T,\Omega,R,h,b^0)$, that identifies modeled factors $X$. Fix the policies $\pi_{-i}$ for the other agents. An \textit{Influence Augmented
		Local Model}
	$\mathcal{M}_i=(\bar{\mathcal{S}}_i,A_i,\mathcal{T}_i,R_i,h,b^0)$ is a
	factored MDP where the action space $A_i$ and rewards $R_i$ correspond to the POSG agent $i$'s actions and rewards. The augmented state space
	$\bar{\mathcal{S}}_i$ consists of local states and d-separating sets as $\bar s^t = (x^t , d^t)=((x^t_{\text{loc}}, x^t_{\text{dest}}),d^t)$.
	The transition functions 
	are derived as
	\begin{align}\label{IALMTransitions}
	&\mathcal{T}_i(\bar s^{t+1}|\bar s^{t},a^t) =  
	T(x_{\text{loc}}^{t+1}|x^{t},a^t)\,\sum_{y_{\textit{src}}^t}T(x_{\text{dest}}^{t+1}|x^{t},y_{\textit{src}}^t,a^t)\,I^t(y_{\textit{src}}^t|d^t)
	\end{align}
\end{definition}
This model defines an equivalent problem to the best-response problem for agent $i$ against $\pi_{-i}$. Namely, the optimal policy for the MDP defined by the IALM $\mathcal{M}_i$ corresponds to the best response against policies $\pi_{-i}$ in the factored POSG $M$. 
The claim and complete proof that IBA provides a lossless abstraction can be found in Section 6 \cite{oliehoek2019sufficient}. 
\subsection{Influence in Planetary Exploration}
\label{InfluenceExample}
To give a concrete intuition of influence-based abstraction we use a version of the planetary exploration environment \cite{witwicki2010influence}. A rover has to explore a planet, and its navigation may be guided by a plan from a satellite. 
The goal of the rover is to move until it reaches a target site, where it will collect a positive reward. However, any time it fails to step forward, it will receive a penalty. The satellite might help the rover by providing a plan which  increases the likelihood of a successful step.

\begin{figure}[h]
	\includegraphics[width=8.6cm]{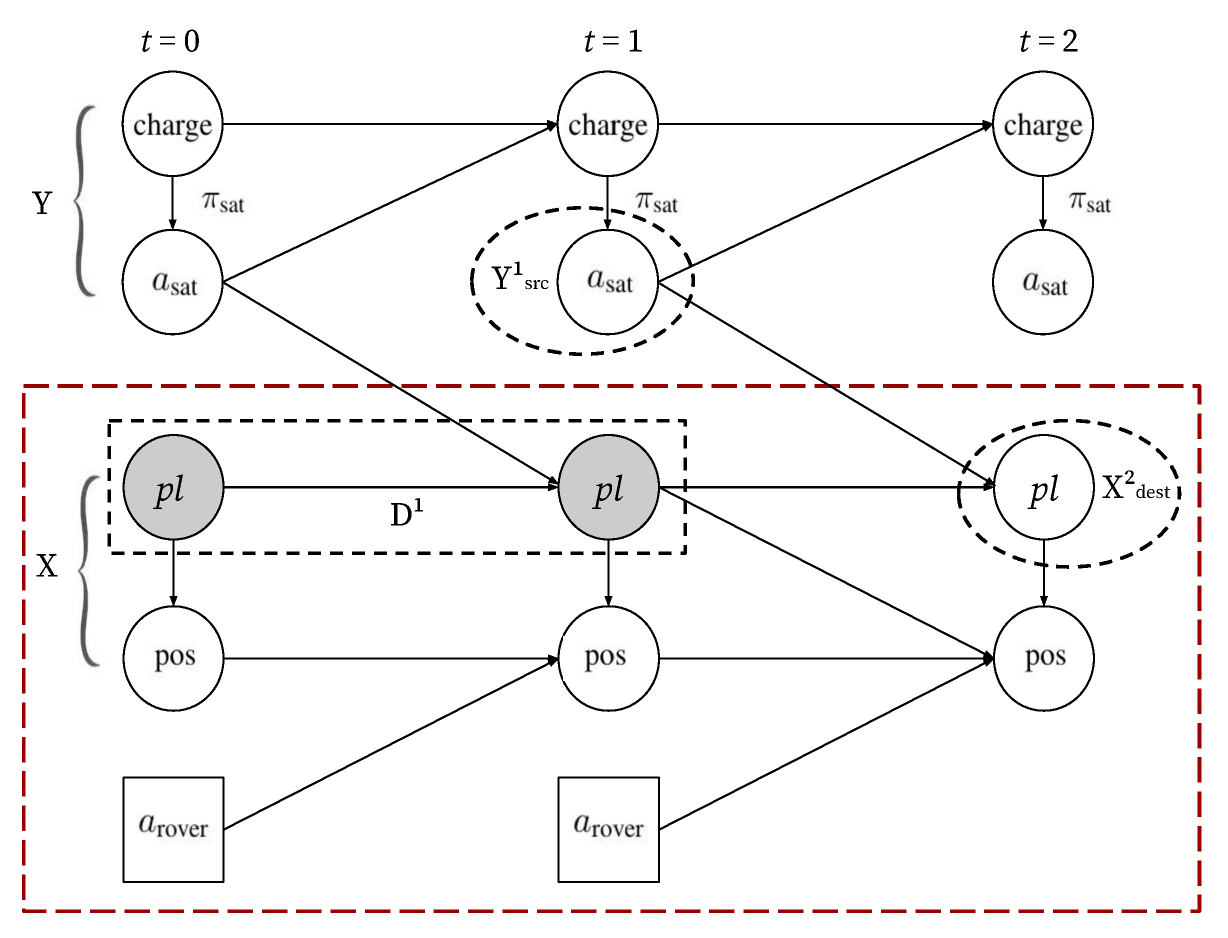}
	\caption{A DBN of the planetary exploration domain. The dotted red square delimits the local model of the rover, including the modeled factors $X$. The external part comprises the non-modeled factors $Y$. Among them, the satellite action $a_\text{sat}$ directly affects the availability of a plan $pl$ and therefore is the influence source. The gray circles constitute the d-set $D^1$ to infer the action $a_\text{sat}$ at time $t=1$. Namely, the history of the plan $(pl^0,pl^1)$ retains the sufficient local information to infer the satellite action $a^1_\text{sat}$. The influence on the local model at the next time step $t=2$ is experienced directly by the modeled factor plan $pl^2$ that is thus the influence destination. 
	}
	\label{roverfig}
	\Description{2-Dynamic Bayesian Network for the planetary exploration domain, showing the dependencies between variables.}
\end{figure}

Figure \ref{roverfig} shows a dynamic Bayesian network (DBN) \cite{boutilier1999decision} that compactly represents the problem. At each time $t$ the rover can observe its position $pos$ and if a plan $pl$ was available at the previous time step.
The rover's local state consists of $X=(pos,pl)$. The plan $pl$ is the influence destination since it is directly affected by the non-modeled satellite action $a_{\text{sat}}$ which is accordingly the influence source.
Contrarily, the position $pos$ is only directly affected by local variables.
The decisions of the satellite might depend on its level of battery, the \textit{charge}, which only indirectly affects the local model through the satellite actions.
Potentially, the decision-making problem of the satellite might depend on a conceivably larger number of variables: it has to manage its own resources, send plans to other rovers etc.
However, the only information the rover needs to retrieve to act optimally is whether the satellite will make available a routing plan at the next step. Therefore, it can abstract away all the other state variables and try to \emph{infer} the satellite's action, given all the relevant information that it has in the local model. In this scenario, it turns out that all it has to remember is the history of the availability of plans at each time step. Therefore the influence point consists of the distribution of the satellite's actions $Y_{\textit{src}}^t=a^t_{\text{sat}}$, given the local history of the plan $D^t=(pl^0,\dots,pl^t)$.

\section{Planning with Approximate Influence Representations}
\label{simbasedplanning}
IBA has the potential to enable faster planning in complex domains where using the entire model would typically be too heavy and computational demanding with no information loss.
However, exact influence computation requires solving a large number of intractable inference problems. In fact, in most applications, the size of the d-set increases linearly in time, resulting in an exponentially increasing number of instantiations of the d-set.
This means that representing an EIP takes exponential space, and computing an EIP requires solving exponentially many and possibly hard inference problems.
This motivates the idea of using \emph{approximate influence point} (AIP) representations: we instead use advances in machine learning that allow us to generalize AIPs over d-separating sets.
The overall idea consists of transforming the global model into a new approximate influence-augmented local model at which any solution method can be applied. Our method is therefore complementary to the planning method. The advantage lies in the reduction of the problem size and therefore in computational complexity.

The approach is straightforward but sketched in Algorithm \ref{AIBA} for completeness: first we collect samples of the d-set $D^t$, the local history sufficient to predict the influence sources $Y^t_{\text{src}}$ in the global model using an exploratory policy $\pi^{\text{Exp}}_i$ for agent $i$ (lines 1-3).
In the second step, we learn the approximate influence point $\hat I^t$ using the $N$ trajectories of $\left (D^t,Y^t_{\text{src}}\right )$ as training set
(lines 4-5).
Then, we construct the approximate local transitions $\hat {\mathcal{T}}_i$ using the learned influence point $\hat I$ according to \ref{IALMTransitions} (line 6). Then, we assume that the resulting local problem might be solved by simulation-based
planning or RL approaches to derive a near optimal best-response $\hat{\pi}^*_i$ for agent $i$ (line 7-9). Note that in line 8 any solution method can be used to solve the MDP problem. The idea is that this phase uses $M \gg N$ episodes, thus making the initial cost of using the expensive global simulations negligible.

While planning with a learned approximate influence could lead to
significant speedups, there is also a risk as the approximate influence
point may induce inaccurate state transitions, leading to a loss in
value. A key question therefore is: what constitutes good AIPs? What
conditions need to hold on $\hat I$, for $\hat{\pi}^*_i$ to be close to optimal?
\begin{algorithm}
	\caption{Exact Best Response to Approximate Influence}
	\label{AIBA}
	\begin{algorithmic}[1]
		\REQUIRE 
		exploratory policy $\pi^{\text{Exp}}_i$, 
		policies $\pi_{-i}$
		\STATE \COMMENT{Simulation:}
		\STATE Create a dataset of $N$ trajectories of the global model
		\STATE Extract $\left (d^t_k,y^t_{\text{src},k}\right )_{k=1:N,t=1:h}$.
		\STATE \COMMENT{Train a Prediction Algorithm:}
		\STATE Learn approximate influence $\hat I^t$ from $\left (d^t_k,y^t_{\text{src},k}\right )_{k=1:N}$ for any $t=0,\dots,h$
		\STATE Compute approximate local transitions $\hat {\mathcal{T}}_i$ from $\hat I$
		\STATE \COMMENT{Solve the local planning/RL problem:}
		\STATE {Compute $\hat{\pi}^*_i=\text{MDPSolver}(\hat {\mathcal{T}}_i)$} \\
		\RETURN $\hat{\pi}^*_i$
	\end{algorithmic}
\end{algorithm}

\section{Theoretical Loss Bounds}
\label{theoreticallossbound}
Here we want to discuss
guarantees on the value loss when using AIPs instead of EIPs to derive best-response policies. Such kind of results allow us to derive conditions for approximate influence to yield near-optimal solutions.

Formally, we consider two IALMs $\mathcal{M}=(\bar{\mathcal{S}},A,\mathcal{T},R,h,b^0)$ and $\hat{\mathcal{M}}=(\bar{\mathcal{S}},A,\hat{\mathcal{T}},R,h,b^0)$ sharing the same augmented state space, action space and rewards. They differ only in the transition functions $\mathcal{T}$ and $\hat{\mathcal{T}}$ that are induced respectively by the EIP $I$ and an AIP $\hat I$ according to \ref{IALMTransitions}. We see $\hat{\mathcal{M}}$ as an approximation to $\mathcal{M}$. We omit the subscript $i$ when referring to an IALM for agent $i$ to ease the notation.


\subsection{$L^1$ Loss Bound}

We introduce first the necessary notation.
We use $\mathcal{V}^*$ to denote the objective optimal value for $\mathcal{M}$. $\hat\pi^*$ refers to the optimal policy in the approximate IALM $\hat{\mathcal{M}}$. $\mathcal{V}^{\hat\pi^*}$ denotes the value achieved by policy $\hat\pi^*$ in the IALM $\mathcal{M}$. Our aim is to bound the value loss given by the difference  $\mathcal{V}^*-\mathcal{V}^{\hat\pi^*}$. We define $|R|\defas\max_{a,s}|R(s,a)|$. We use $||\,.\,||_1$ and $D_{KL}$ to denote the $1$-norm and the KL divergence between probability distributions.

The proof of the value loss bound derived is divided into two steps. We proceed first by bounding the value loss with the difference between the IALM transitions functions $||\mathcal{T}-\hat{\mathcal{T}}||_1$. Then, we prove an upper bound for $||\mathcal{T}-\hat{\mathcal{T}}||_1$ in terms of the difference  $||I-\hat I||_1$. Finally, combining these two results we derive an upper bound for the value loss in terms of the distance between the EIP and the AIP $||I-\hat I||_1$.
We collect the proofs of all the following claims in the Appendix A.
%

Since the IALM $\hat{\mathcal{M}}$ is essentially an approximation of the exact influence IALM $\mathcal{M}$ with imprecise transitions, we apply the loss bounds for MDPs with uncertain transition probabilities \cite{delgado2011efficient,strehl2009reinforcement, mastin2012loss} to the case of the IALMs.
\begin{theorem}\label{ThTransitions}
	Consider the two IALMs $\mathcal{M}$ and $\hat {\mathcal{M}}$, the following loss bound holds:
	\begin{equation}\label{TransitionBoundsEq}
	||\mathcal{V}^*-\mathcal{V}^{\hat\pi^*}||_{\infty}\leq 2h^2|R|\, \max_{t}\max_{\bar s^t,a^t}||(\mathcal{T}(\cdot|\bar s^t,a^t)- \hat {\mathcal{T}}(\cdot|\bar s^t,a^t)||_{1}
	\end{equation}
\end{theorem}

The importance is that we can bound the value loss when the IALM transition functions
$\mathcal{T},~\hat{\mathcal{T}}$ induced by $I,~\hat{I}$ are sufficiently close.

For the second step, we use the relation between the IALM transitions $\mathcal{T}$ and the influence point $I$ expressed by \ref{IALMTransitions}.
\begin{lemma}\label{TransitionsVSInfluence}
	Consider the IALMs transitions $\mathcal{T}$ and $\hat {\mathcal{T}}$. For any $t$, augmented state $\bar s^t=(x^t,d^{t})$ and action $a^{t}$,
	\begin{align}\label{TransitionsVSInfluenceEQ}
	\begin{split}
	&||(\mathcal{T}(\cdot|\bar s^t,a^t)- \hat {\mathcal{T}}(\cdot|\bar s^t,a^t)||_{1}\leq ||(I^t(\cdot|\,d^t)- \hat {I}^t(\cdot|\,d^t)||_{1}
	\end{split}
	\end{align}
\end{lemma}
Combining Theorem \ref{ThTransitions} and Lemma \ref{TransitionsVSInfluence}, we obtain the $L^1$ loss bound.
\begin{theorem}\label{ValueVSInfluence}
	Consider an IALM $\mathcal{M}=(\mathcal{S},A,\mathcal{T},R,h,b^0)$ and an AIP $\hat I$ inducing $\hat{\mathcal{M}}=(\mathcal{S},A,\hat{\mathcal{T}},R,h,b^0)$.
	Then, a value loss bound in terms of the $1$-norm error is given by
	\begin{equation}\label{InfluenceBoundEq}
	||\mathcal{V}^*-\mathcal{V}^{\hat\pi^*}||_{\infty}\leq 2h^2 |R| \max_{t,\,d^{t}} ||(I^t(\cdot|\,d^t)- \hat {I}^t(\cdot|\,d^t)||_{1}
	\end{equation}
\end{theorem}

This shows that the value loss is bounded by the worst case $L^1$-distance between the exact and the approximate influence over all the possible d-set instantiations.
\subsection{KL Divergence Bound}
We now present a loss bound in terms of the KL divergence between the approximate and the exact influence. Since the cross entropy and KL divergence differ solely by an additive constant, this result establishes a relation between the value loss and the cross entropy error for influence approximation. 
\begin{corollary}\label{ValueVSInfluenceKL}
	Consider an IALM $\mathcal{M}=(\mathcal{S},A,\mathcal{T},R,h,b^0)$ and an AIP $\hat I$ inducing $\hat{\mathcal{M}}=(\mathcal{S},A,\hat{\mathcal{T}},R,h,b^0)$.
	Then, a value loss bound in terms of KL divergence error is given by
	\begin{equation}\label{InfluenceBoundEqKL}
	||\mathcal{V}^*-\mathcal{V}^{\hat\pi^*}||_{\infty}\leq 2h^2 |R| \max_{t,d^{t}} \sqrt{2 D_{KL}(I^t(\cdot|\,d^t)||\hat I^t(\cdot|\,d^t))}
	\end{equation}
\end{corollary}
We see that the performance loss can be bounded by the max KL divergence.
Cross entropy loss optimizes the mean of such loss $\mathbb{E}_{t,D^t}\left [ D_{KL}(I^t(\cdot|D^t))||\hat I^t(\cdot|D^t)\right ]$ and therefore is aligned with this bound: even though it does not optimize the max itself, it is
intuitively clear that in many problems a low mean will imply a low max error, even the more since the mean is known to be sensitive to outliers. Moreover, this is an asymptotically tight bound. That is, when the KL divergence tends to zero the approximate influence solution approaches the true optimal one.
This suggests that neural networks learning approaches are well suited for the approximate influence learning task and that the cross entropy, which has been widely used as test error, can give a priori insight on the value loss.
\subsection{Probabilistic Loss Bound}
\label{probabilisticbounds}
The previous section gives an idea of what AIPs properties can guarantee bounded value loss. However, the direct application of Theorem~\ref{ValueVSInfluence} and Corollary~\ref{ValueVSInfluenceKL} requires the distance to the true influence point $I$, which is unknown.
In this section, we present an approach to overcome this limitation:
a probabilistic loss bound depending only on the distance to an empirical influence
distribution, which can be measured using a test set.

Precisely, assume that for a given instatiation $d^t$ of the d-set at time $t$ we have $N$ samples of the influence sources $\left \{y^t_{\text{src},1},\dots,y^t_{\text{src},N}\right \}$ and let  
$$
I^t_N\,(\,y^t_{\text{src}}\, |\, d^{t})=\frac{1}{N}\sum_{k=1}^N \delta_{y^t_{\text{src},k}}(y^t_{\text{src}})
$$
denote the empirical conditional distribution.
Our result presents a value loss bound in terms of the distance between the empirical influence $I^t_N$ and the approximate $\hat I^t$.
\begin{theorem}\label{ProbBound}
	Consider an IALM $\mathcal{M}=(\mathcal{S},A,\mathcal{T},R,h,b^0)$ and an AIP $\hat I$ inducing $\hat{\mathcal{M}}=(\mathcal{S},A,\hat{\mathcal{T}},R,h,b^0)$. Assume that for every time $t$ and d-set instantiation $d^t$, we have at least an sample of size $N$ of influence sources.
	Then  for every $\varepsilon >0$
	\begin{align} \label{probboundeq}
	&\mathbb{P}\left (||\mathcal{V}^*-\mathcal{V}^{\hat\pi^*}||_\infty\leq 2 h^2 |R|\max_{t,d^t}||(I^t_N(\cdot|\,d^t)- \hat {I}^t(\cdot|\,d^t)||_{1}+\varepsilon\right )\geq\nonumber\\ &1-h\,|D^h|(2^{|Y_{\text{src}}|}-2)e^{-N\frac{\varepsilon^2}{2}}
	\end{align}
	where $|Y_{\text{src}}|$ is the cardinality of the influence sources space and $|D^h|$ the cardinality of the d-sets space at time $h$.
\end{theorem}
This theorem states that, with high probability, the value loss of using an AIP $\hat I$ is upper bounded by the empirical error (that we can measure using a test set).
We believe that this is a first step towards bounds that can be used in practical applications.

However, the bound in Theorem~\ref{ProbBound}, still requires an intractable maximization over all possible instantiations of d-sets (and needs $N > \log(h|D^h|2^{|Y_{\text{src}}|})$ samples for each of these). Moreover, the maximization over d-sets might be too conservative in many cases. Namely,
the right-hand side of ~(\ref{InfluenceBoundEqKL}) corresponds to the loss one agent would incur if every time step the mistake is caused by the worst-case approximation of the influence over d-sets. We think that the maximization may be replaced with an appropriate expectation over instantiations. In this respect, identifying versions of these bounds
that work with expectations over d-sets, rather than maximization, is one of the important directions of future work that our paper identifies. 

\section{Multiagent Implications}
\label{Multiagentpersp}
So far, we have focused on the perspective of a single agent, that computes a best
response against the fixed policies of other agents. However, the results in this paper
can have important implications also for settings where we optimize or solve for the
policies of multiple agents at the same time. In fact, the insights of influence-based
abstraction originated from the study of multiagent systems
\cite{becker2004decentralized,
	becker2004solving,
	varakantham2009exploiting,
	Petrik09JAIR,
	Witwicki09AAMAS,
	Witwicki10AAMAS,
	witwicki2010influence,
	Velagapudi11AAMAS,
	Witwicki11AAMAS,
	Witwicki12AAMAS,
	Oliehoek15IJCAI},
so these implications should hardly be surprising. Nevertheless, we think it is useful to
spell out some of these implications and will do so in the remainder of this section.

First, we point out that the best-response setting that we consider is very general. In
fact, the notion of a `fixed policy' of an agent in a POSG is very powerful: such a fixed
policy is a mapping from histories of actions and observations to distributions over
actions and therefore powerful enough to model learning agents. This means that the fixed
policies that we compute a best response against could include, e.g., a Q-learning agent.

A second implication is for the computation of equilibria in POSGs. Many methods for
computing Nash equilibria in POSGs
\cite{berger2007brown,brown1951iterative,mcmahan2003planning,oliehoek2006parallel,Lanctot17NIPS}, use
best-response computation as an inner loop. In many cases, such as in the double-oracle
algorithm \cite{mcmahan2003planning} or the parallel Nash Memory
\cite{oliehoek2006parallel}, replacing such a best-response
computation with an $\epsilon$-best response computation can enable us to compute
$\epsilon$-approximate Nash Equilibria ($\epsilon$-NEs) \cite{Nisan07bookAGT}, in which no player can
benefit more than $\epsilon$ from deviating. As such, our results -- that show under what
conditions on the influence predictions we can compute an $\epsilon$-approximate best response -- may lead to
computationally feasible paths to compute $\epsilon$-NEs.

Even in cases where computing $\epsilon$-NEs will remain out of reach, our results can
provide insight in complex MAS. 
\begin{observation}
	Given a method $E$ to estimate an AIP with an error (maximum $L_1$ norm) of
	at most $\epsilon_1$, we can verify if a particular joint policy $\pi=(\pi_1,\dots,\pi_n)$ is an $(2h^2|R|\epsilon_1)$-NE as
	follows:
	1) use $E$ to estimate the influence $\hat I_i$ on each agent $i$
	2) verifying if $\pi_i$ is an optimal solution in the IALM constructed for agent~$i$.
\end{observation}
Of course, the question of how to develop such estimators $E$ is not trivial in the
general case, but there are many special cases with compact influence descriptions where this is possible
\cite{oliehoek2019sufficient,Chitnis20CORL}.

Moreover, IBA serves as the basis for \emph{influence search} \cite{becker2004solving,witwicki2010influence,Witwicki12AAMAS} in cooperative settings. 
The key idea is that an influence point captures all the relevant information about many
different policies of the other agents. That is, different joint policies might induce the
same influences on the local models of the agents. Thus, the space of `joint influence
points' $I=(I_1,\dots,I_n)$ can be much smaller than the space of joint policies
$\pi=(\pi_1,\dots,\pi_n)$, 
and it can be much more efficient to search through the former.
\cite{Witwicki11AAMAS}.\footnote{
	Roughly speaking, an influence search algorithm searches 
	in the space of possible joint influences, and for each joint influence point
	$I=(I_1,\dots,I_n)$, it solves a local constrained best-response problem for each
	agent (possibly in parallel): each agent $i$ computes a best-response $\hat\pi^*(I)$
	to $I_i$ constrained to inducing influences $I_{-i}$ on the
	local models of the other agents. 
}
\section{Empirical evaluation}
\label{experiments}
\begin{figure*}[h]
	\begin{subfigure}{0.4\textwidth}
		\includegraphics[scale=0.47]{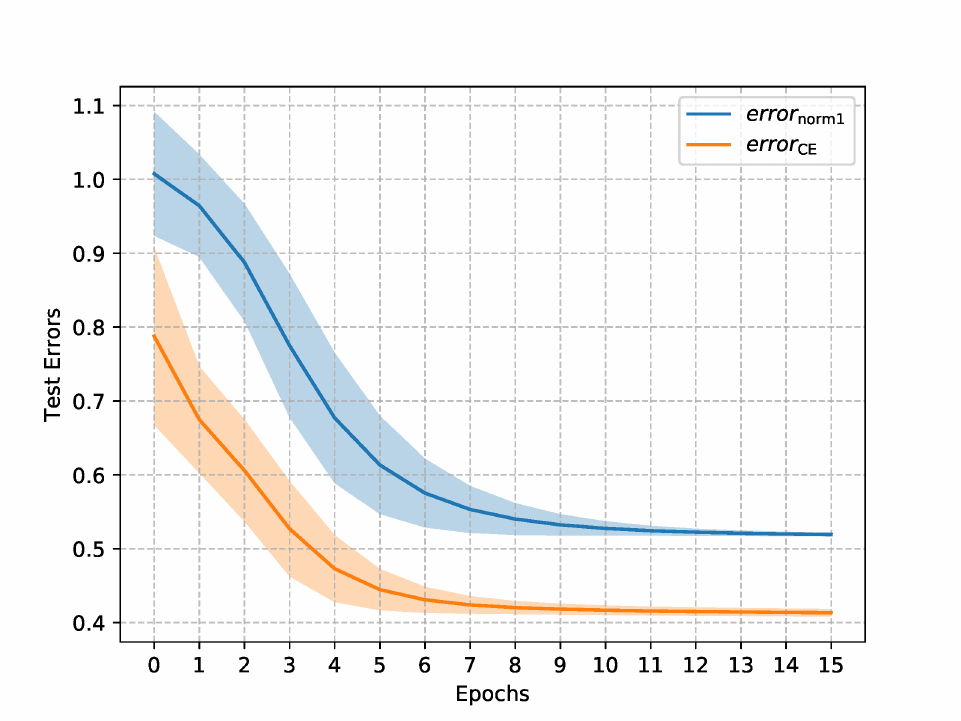}
		\caption{Test errors for increasing training epochs. \\ \mbox{ }}
		\label{TestErrMR}
		\Description{Plot of cross entropy and L1 error (y-axis) as functions of the number of training epochs (x-axis) for the Planetary Exploration domain. The figure shows both errors as decreasing functions in the number of epochs.}
	\end{subfigure}
	\begin{subfigure}{0.4\textwidth}
		\includegraphics[scale=0.47]{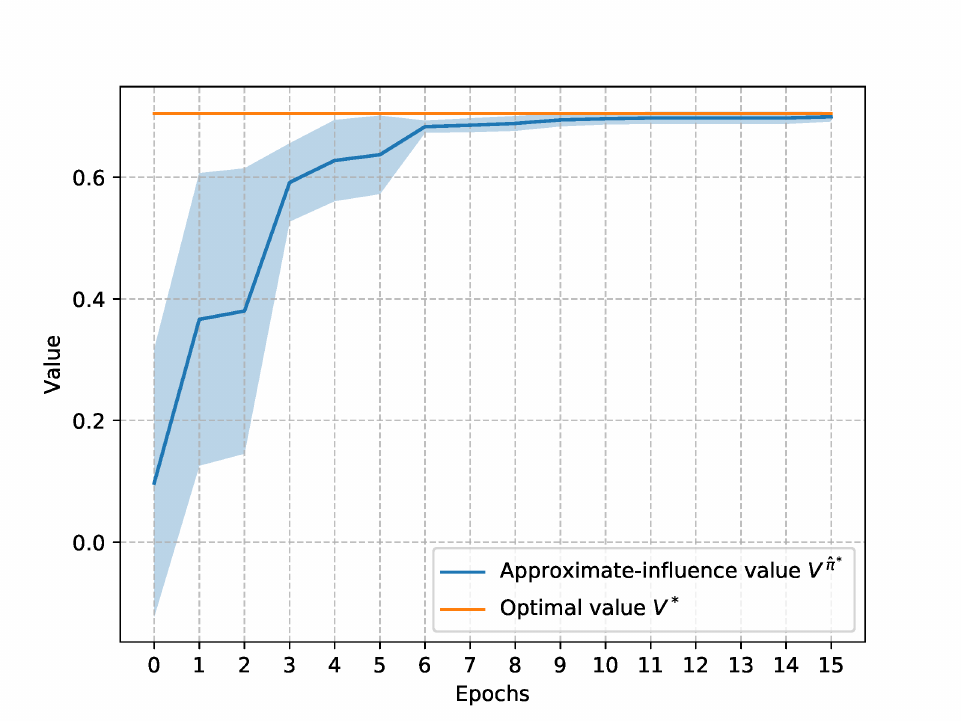}
		\caption{Value achieved by $\hat{\pi}^*$ compared to the optimal value for increasing training epochs.}
		\label{ValueMR}
		\Description{Plot of the value achieved by the approximate-influence optimal policy (y-axis) and the optimal value as functions of the training epochs (x-axis) for the Planetary Exploration domain. The figure shows that the value for the approximate-influence increases in the number of epochs.}
	\end{subfigure}
	\hspace{-0.8cm}
	\begin{subfigure}{0.4\textwidth}
		\includegraphics[scale=0.47]{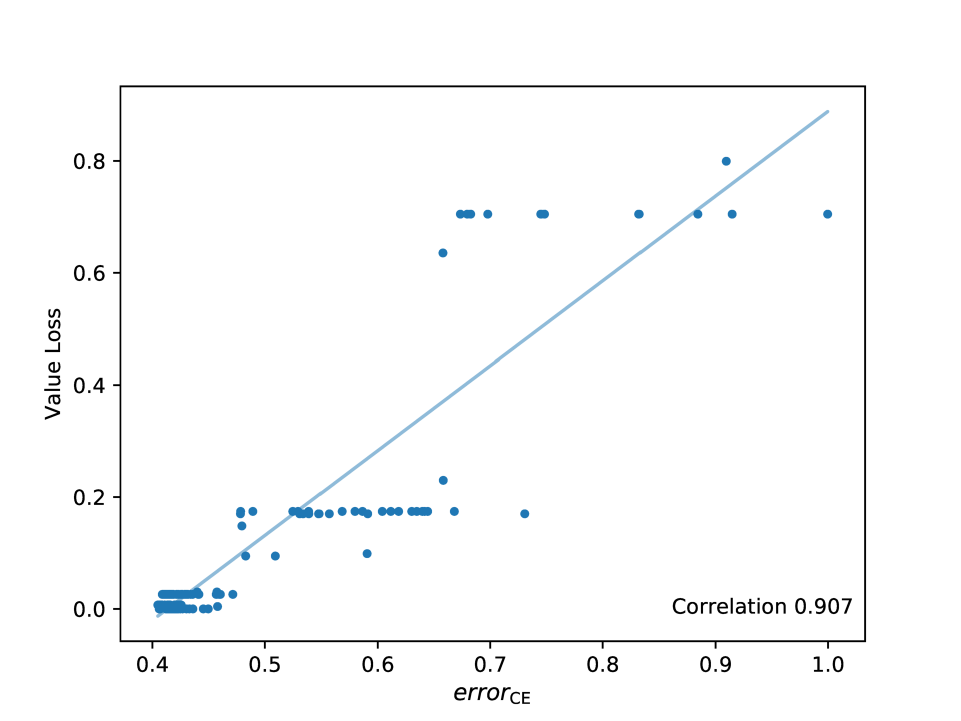}
		\caption{Correlation between value loss and $error_{\text{CE}}$.}
		\label{CorrCE_MR}
		\Description{Plot of value loss (y-axis) and the corresponding cross entropy error (x-axis) for the Planetary Exploration domain. The figure shows that the correlation between these two variables is around 0.9.}
	\end{subfigure}
	\begin{subfigure}{0.4\textwidth}
		\includegraphics[scale=0.47]{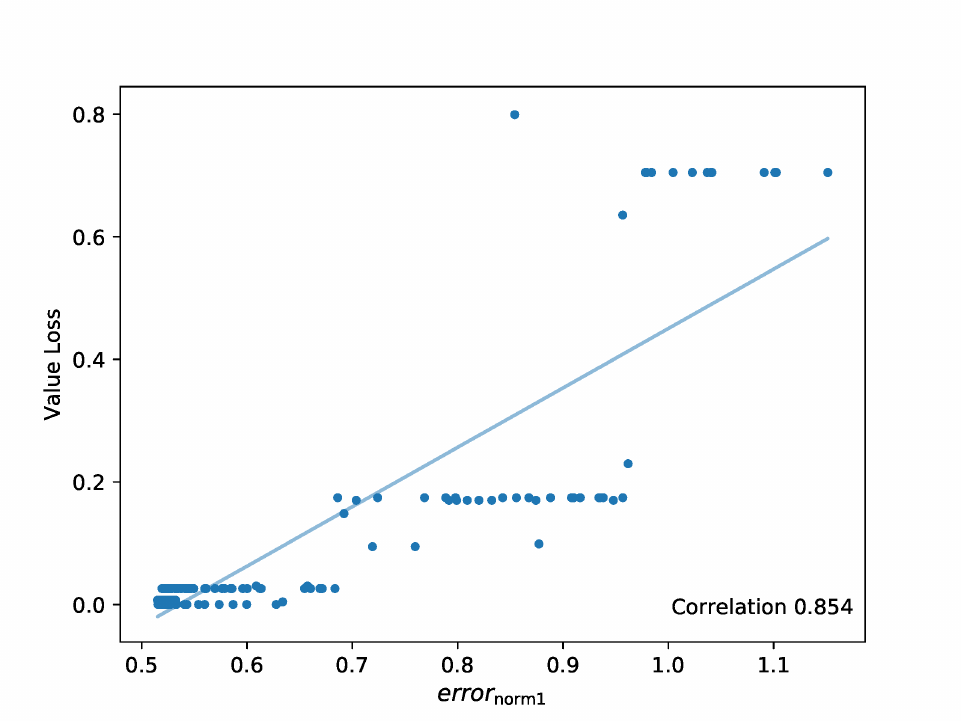}
		\caption{Correlation between value loss and $error_{\text{norm1}}$.}
		\label{CorrN1_MR}
		\Description{Plot of value loss (y-axis) and the corresponding norm 1 error (x-axis) for the Planetary Exploration domain. The figure shows that the correlation between these two variables is around 0.85.}
	\end{subfigure}
	\caption{Planetary Exploration.} 
	\label{MR}
\end{figure*}
The bounds presented in Section \ref{theoreticallossbound}
establish a relation between the value loss and the cross entropy of the influence approximations. In particular, they suggest that the \emph{mean} cross entropy loss is aligned with the objective of minimizing the performance loss.

Here we evaluate if this alignment translates into practical settings by investigating the relation between the mean cross entropy 
$$\text{CE}(I,\hat I)\triangleq\mathbb{E}_{t,D^t}\left [\text{ CE}(I^t(\cdot|D^t),\hat I^t(\cdot|D^t))\right ]$$
and the performance loss $\mathcal{V}^*-\mathcal{V}^{\hat\pi^*}$. Namely, we expect to see that a better influence approximation in terms of cross entropy loss, indeed, leads to a smaller value loss. Here we try to validate this hypothesis.

Similarly, we investigate to what extent the \emph{mean} $1$-norm error 
$$||I-\hat I||_1\triangleq\mathbb{E}_{t,D^t}\left [||I^t(\cdot|D^t)-\hat I^t(\cdot|D^t)||_1\right ]$$
allows to a priori assess the quality of the approximations in terms of the value loss. In fact, according to the results in Section \ref{theoreticallossbound}, the $1$-norm has the potential to provide a tighter bound for the performance loss. If experimentally confirmed, this would motivate future work to investigate the possibility to use different training losses based on $L^1$ distance.

\begin{figure}[h]
	\includegraphics[scale=0.37]{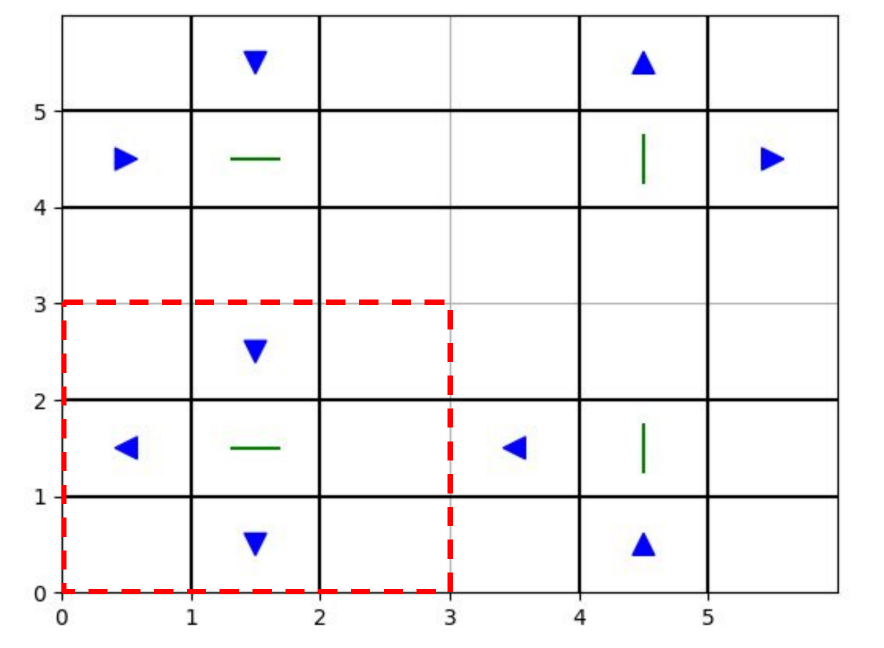}
	\caption{Traffic network. The local model is delimited by the dotted square. The blue triangles represent the vehicles and the green bars the traffic lights.
	}
	\label{TrafficGrid}
	\Description{Depiction of one state of the Markov decision process representing the traffic network domain. It consists of 4 road segments (2 horizontal and 2 vertical) which cross forming a regular grid. The cars are represented by blue triangles which indicate the direction of travel of each road segment. At each intersection, a red bar represent a traffic light.}
\end{figure}
\begin{figure*}[h]
	\begin{subfigure}{0.4\textwidth}
		\includegraphics[scale=0.47]{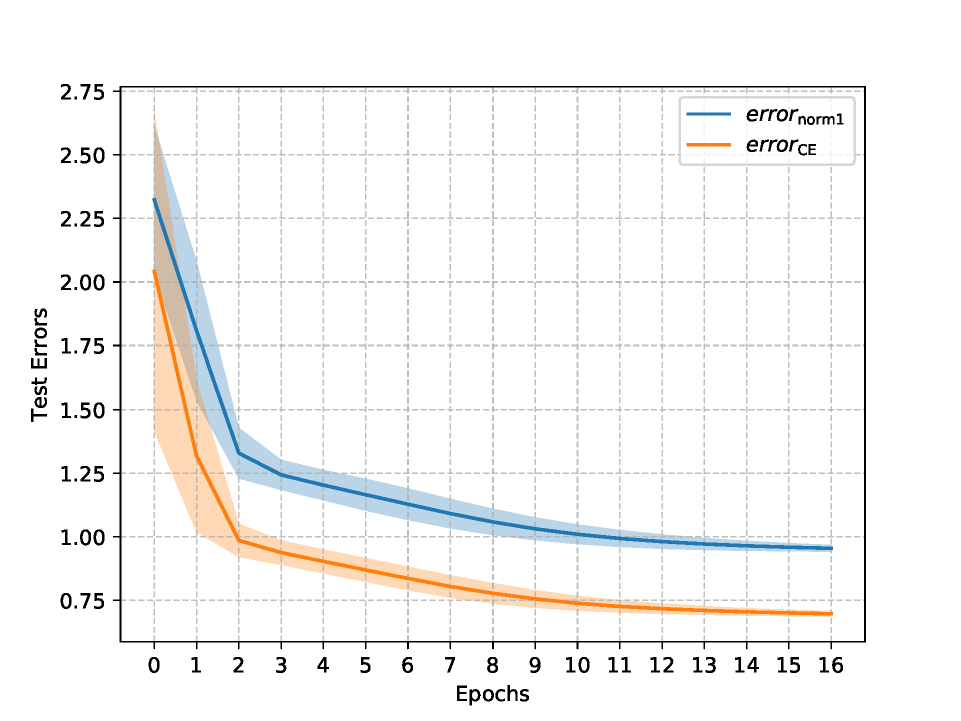}
		\caption{Test errors for increasing training epochs. \\ \mbox{ }}
		\label{TestErrTD}
		\Description{Plot of cross entropy and norm 1 error (y-axis) as functions of the number of training epochs (x-axis) for the Traffic domain. The figure shows both errors as decreasing functions in the number of epochs.}
	\end{subfigure}
	\begin{subfigure}{0.4\textwidth}
		\includegraphics[scale=0.47]{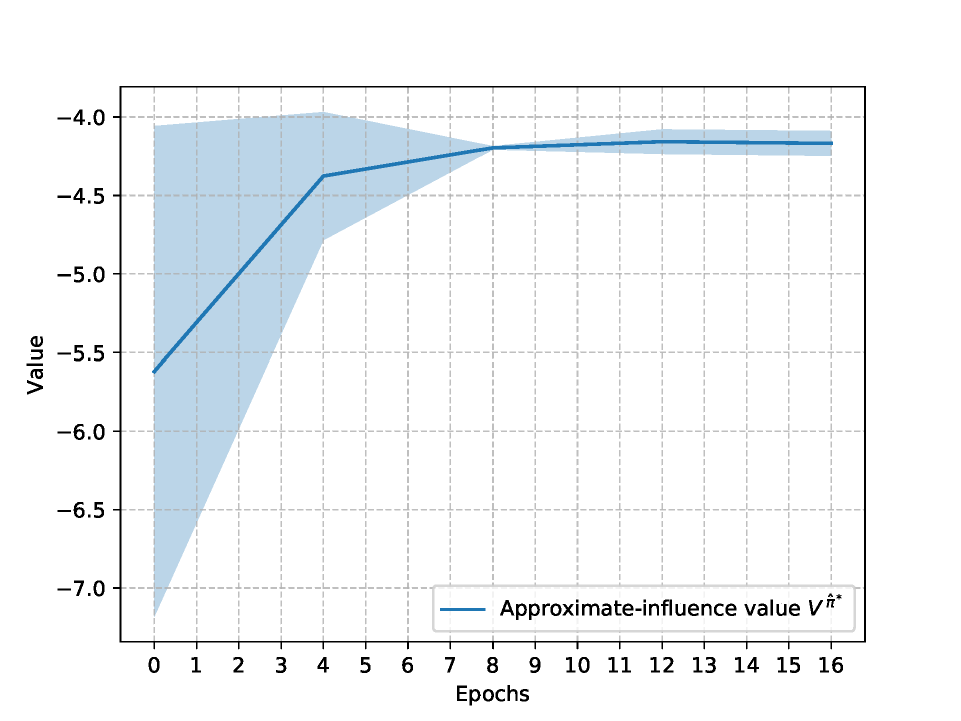}
		\caption{Value achieved by $\hat{\pi}^*$ for increasing training epochs computed every $4$ epochs.}
		\label{ValueTD}
		\Description{Plot of the value achieved by the approximate-influence optimal policy (y-axis) as a function of the training epochs (x-axis) for the Traffic domain. The figure shows that the value for the approximate-influence increases in the number of epochs.}
	\end{subfigure}
	\hspace{-0.8cm}
	\begin{subfigure}{0.4\textwidth}
		\includegraphics[scale=0.47]{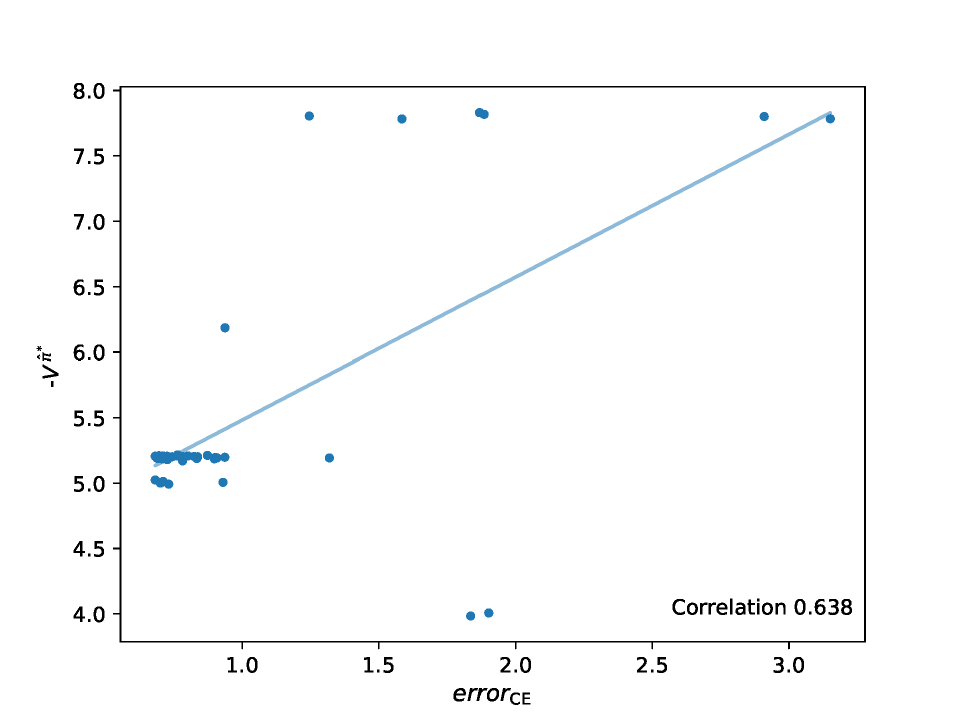}
		\caption{Correlation between value loss and $error_{\text{CE}}$.}
		\label{CorrCE_TD}
		\Description{Plot of value loss (y-axis) and the corresponding cross entropy error (x-axis) for the Traffic domain. The figure shows that the correlation between these two variables is around 0.64.}
	\end{subfigure}
	\begin{subfigure}{0.4\textwidth}
		\includegraphics[scale=0.47]{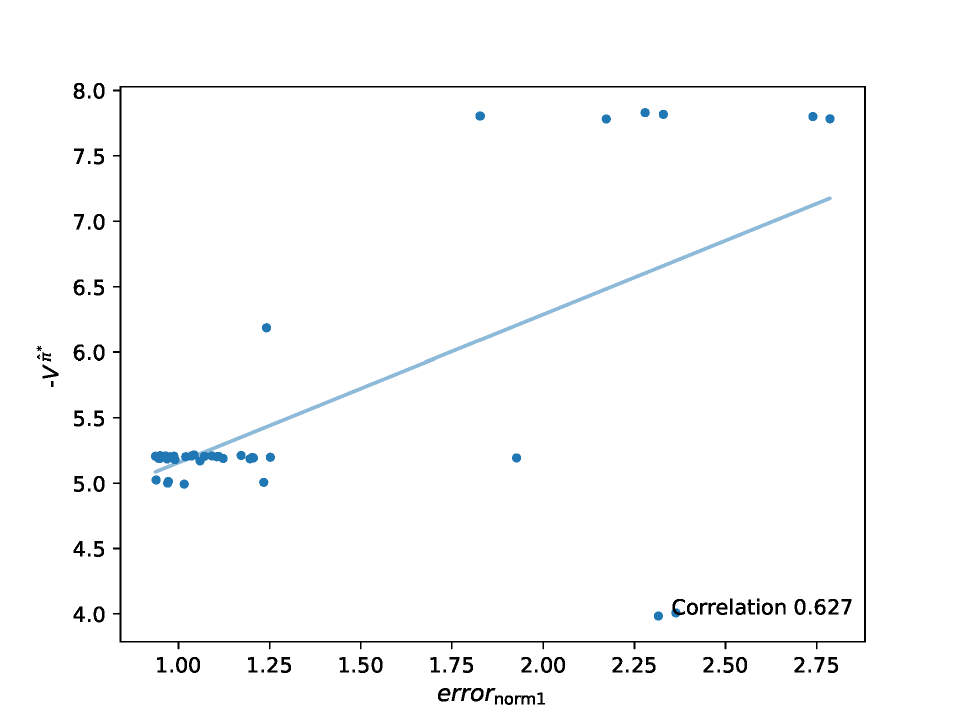}
		\caption{Correlation between value loss and $error_{\text{norm1}}$.}
		\label{CorrN1_TD}
		\Description{Plot of value loss (y-axis) and the corresponding norm 1 error (x-axis) for the Traffic domain. The figure shows that the correlation between these two variables is around 0.63.}
	\end{subfigure}
	\caption{Traffic network.} 
	\label{TD}
\end{figure*}
\subsection{Experimental Setup}
We follow the procedure sketched in Algorithm \ref{AIBA}.
That is, we first run the simulations from the global model using an exploratory random policy $\pi^{\text{Exp}}$ for the local agent and the set of fixed policies for the other agents. We collect the samples of influence sources and d-sets $\{y^t_{\text{src,k}},d^t_k\}_{t=1:h,k=1:N}$. We train a LSTM neural network with cross entropy loss to induce the approximate influence $\hat I^t(\,\cdot\,| D^t)$. At different training epochs, for the corresponding approximations of the influence, we build the approximate-influence local model. Then, we compute the optimal policy $\hat \pi^*$ in the approximate-influence model through value iteration \cite{bertsekas1995dynamic}. We evaluate the policy $\hat \pi^*$ in the global model using $M$ simulations to obtain the value achieved $\mathcal{V}^{\hat\pi^*}$. 
Note that since we use an exact solution method to compute the policy $\hat\pi^*$ in the approximate-influence model, any other solution method would get lower performance in the local model.
For each epoch, to assess the quality of the approximation, we use Monte Carlo estimators for the mean cross entropy $\text{CE}(I,\hat I)$ and the $1$-norm error $||I-\hat I||_1$, computed using a test set as
\begin{align}\label{EmpCeN1}
error_{\text{CE}} &=-\frac{1}{h}\sum_{t=1}^h\frac{1}{N}\sum_{k=1}^N \ln\left (\hat{I}^t(y^t_{\text{src,k}} \mid d^t_k)\right )\nonumber\\
error_{\text{norm1}}& = \frac{1}{h}\sum_{t=1}^h\frac{1}{N}\sum_{k=1}^N \, \left  \lVert \delta_{y^t_{\text{src,k}}}(\,\cdot\,)-\hat I^t(\,\cdot\,\mid d^t_k)\,\right \rVert_1
\end{align}
We compute the average and standard deviation of the $error_{\text{CE}}$ and $error_{\text{norm1}}$ over different iterations of the same experiment. For any iteration, we repeat entirely the steps described above: we recollect training and testing samples from the global model, retrain the neural network etc.
When the problem is sufficiently easy to solve, we compute the optimal value $\mathcal{V}^*$ and then we measure the correlation between $error_{\text{CE}}$, $error_{\text{norm1}}$ and the value loss $\mathcal{V}^*-\mathcal{V}^{\hat\pi^*}$.
Otherwise, we measure the correlation between the test errors and $-\mathcal{V}^{\hat\pi^*}$. In fact, since the optimal value $\mathcal{V}^*$ is constant for increasing epochs, it does not affect the correlation. That is, $Corr(error,\mathcal{V}^*-\mathcal{V}^{\hat\pi^*})=Corr(error,-\mathcal{V}^{\hat\pi^*})$.

We run the experiments in three domains: a version of the planetary exploration \cite{witwicki2010influence}, a traffic domain and the fire fighters problem \cite{oliehoek2010value}. 
Section \ref{InfluenceExample}  provides the overview of the planetary exploration domain and high level descriptions of the other domains follow. For more details we refer to the Appendix B. Our codebase is available at \url{https://github.com/INFLUENCEorg/Approx-IBA-Planning}.
\subsubsection{Traffic Network}
In this domain, we simulate a traffic network with $4$ intersections (see Figure \ref{TrafficGrid}).
The sensors of the traffic lights at each intersection provide information on the $3\times 3$ local grid around them. In Figure \ref{TrafficGrid}, the dotted red square represents the local model for the protagonist agent. The other traffic lights employ hand-coded policies prioritizing fixed lanes. The goal of the agent is to minimize the total number of vehicles waiting at the local intersection. In order to act optimally, the local agent only needs to predict if there will be incoming cars from the $east$ and $north$ lanes of the local model the next time step.
The outgoing cars have some probability to re-enter in the network from other lanes. That is, the outgoing vehicles from $south$ and $west$ can affect the decisions of other traffic lights and consequently, the vehicles inflow in the local model at future time steps. 
\subsubsection{Fire fighters}
We model a team of $2$ agents that need to cooperate to extinguish fires in a row of $3$ houses. At every time step, every agent can choose to fight fires at one of its $2$ neighboring houses. The goal of each agent is to minimize the number of neighboring houses that are burning. We take the perspective of one agent with a local model including only the two neighboring houses. 

\subsection{Experimental Results}
\begin{figure}[h]
	\centering
	\begin{subfigure}{0.4\textwidth}
		\centering
		\includegraphics[scale=0.42]{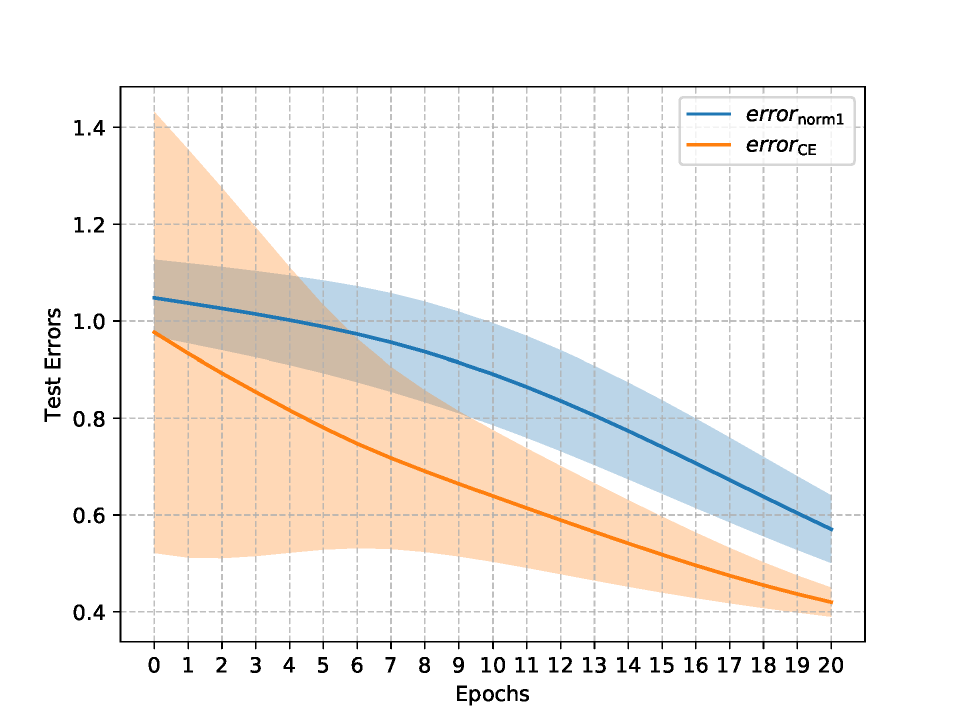}
		\caption{Test errors for increasing training epochs.}
		\label{TestErrFF}
		\Description{Plot of cross entropy and norm 1 error (y-axis) as functions of the number of training epochs (x-axis) for the Fire Fighters domain. The figure shows both errors as decreasing functions in the number of epochs.}
	\end{subfigure}\vfill
	\begin{subfigure}{0.4\textwidth}
		\centering
		\includegraphics[scale=0.42]{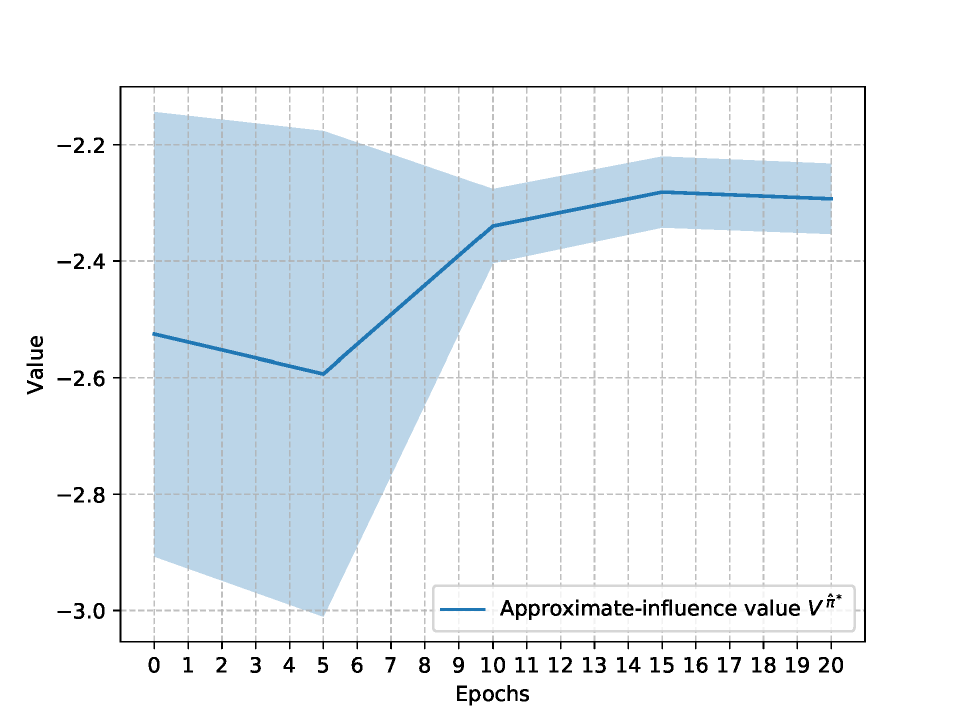}
		\caption{Value achieved by $\hat{\pi}^*$ for increasing training epochs computed every $5$ epochs.}
		\label{ValueFF}
		\Description{Plot of the value achieved by the approximate-influence optimal policy (y-axis) as a function of the training epochs (x-axis) for the Fire Fighters domain. The figure shows that the value for the approximate-influence increases in the number of epochs.}
	\end{subfigure}
	\centering
	\caption{	
		Fire Fighters Problem.}
	\label{FF}%
\end{figure}

Figures \ref{TestErrMR}, \ref{TestErrTD}, \ref{TestErrFF} show the test errors measured by $error_{\text{CE}}$ and $error_{\text{norm1}}$ \ref{EmpCeN1}, as functions of the training epochs for the three domains.
They show that performance in terms of cross entropy and $1$-norm test errors improves monotonically with the number of epochs. These trends suggest that the AIP improves with training, the question is if this also corresponds to a value loss decrease. Looking at Figure \ref{ValueMR} this seems to be the case in the planetary exploration setting. We see that the performance of the policy derived from the approximate-influence IALM is very close to optimal from 6 epochs onward, right about when the training of the neural network starts to stagnate. Moreover, Figure \ref{CorrCE_MR} and \ref{CorrN1_MR} show that the decrease in mean empirical cross entropy or $1$-norm error indeed correlates well with actual value loss.
In the traffic domain, we also see that there is a significant improvement in the value achieved by the approximate-influence policy $\hat\pi^*$ for increasing number of epochs in Figure \ref{ValueTD}. The test errors seem still well aligned with the value loss in Figures \ref{CorrCE_TD}, \ref{CorrN1_TD}.
For the fire fighters domain, Figure \ref{ValueFF} shows that the value improves over training epochs coherently with the decrease of the test errors.

For more results on correlation analysis on different settings of the planetary exploration, see Appendix C.

We can conclude that both the empirical cross entropy error and the $1$-norm errors correlate well with the value loss and therefore provide a priori insight on the quality of the influence-approximation in terms of the value achieved. 

\section{Conclusions and Discussion}
\label{conclusion}
In this paper, we provided a priori quality guarantees for the value loss resulting from approximating the influence point. Our results show that the objective of a neural network trained with the cross entropy loss is well aligned with the goal of minimizing the value loss. Our empirical results demonstrate that empirical (mean) estimators of the cross entropy and the  $1$-norm are predictive of the value loss. We used this insight to evaluate the transfer of our theoretical results to practical scenarios.

In this work, we limit the discussion to fully observable local models. However, in the general case, we would be dealing with a local POMDP. We believe that a generalization of our bound in the partially observable case can be obtained for instance leveraging the `back-projected value vectors' formulation \cite{Oliehoek15IJCAI}. As future work, we intend to further investigate this.

The current formulation of the loss bounds involves a maximization over the space of all the possible instantiations of the d-set, which might grow exponentially in time. In future work, we would like to derive expressions that do not need a maximization, but for example are in expectation over a sampling process. This would lead to tighter bounds, which could be used in practice for model selection or value loss estimation.

\begin{acks}
	This project had received funding from the European Research Council (ERC) under the European Union's Horizon 2020 research and innovation programme (grant agreement No.758824 \textemdash INFLUENCE).
	\\
	\begin{center} \includegraphics[width=0.4\columnwidth]{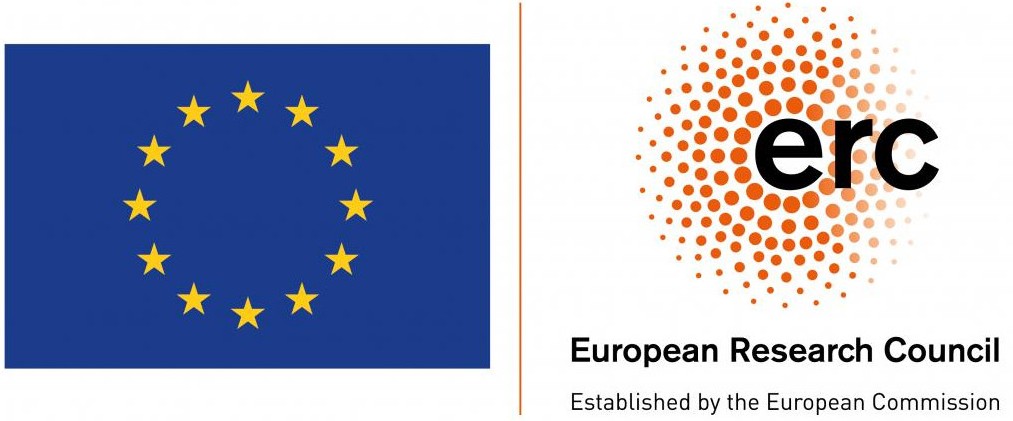}
	\end{center}
\end{acks}



\clearpage
\bibliographystyle{ACM-Reference-Format} 
\balance
\bibliography{bib}
\onecolumn
\appendix
\begin{flushleft}
\section{Theoretical Loss Bounds proofs}
Here we collect all the proofs of the claims in Section \ref{theoreticallossbound}. We use the same notation introduced in Section \ref{theoreticallossbound}.
\begin{customthm}{1}
	Consider the two IALMs $\mathcal{M}$ and $\hat {\mathcal{M}}$, the following loss bound holds:
	\begin{equation*}
	||\mathcal{V}^*-\mathcal{V}^{\hat\pi^*}||_{\infty}\leq 2h^2|R|\, \max_{t}\max_{\bar s^t,a^t}||(\mathcal{T}(\cdot\mid\bar s^t,a^t)- \hat {\mathcal{T}}(\cdot\mid\bar s^t,a^t)||_{1}.
	\end{equation*}
	\begin{proof}
		We apply Corollary 2 \cite{mastin2012loss} to the case of uncertainty in transition probabilities for IALMs. More precisely, Corollary 2 states that given a finite-horizon MDP $M=(S,A, R,T,h)$ and an approximate MDP with uncertain transitions $\hat M=(S,A, R, \hat T,h)$, the performance loss when using the optimal policy $\hat \pi^*$ for the approximate transition model $\hat M$ in the true model $M$ is bounded by
		$$
		||V^*-V^{\hat\pi^*}||_{\infty}\leq 2 h^2 |R|\,\max_{s,a}||T(\cdot\mid s,a)-\hat T(\cdot\mid s,a)||_1.
		$$
		To apply this result to the IALMs $\mathcal{M}$ and $\hat {\mathcal{M}}$, it suffices to maximize the right-hand side of the above expression over time. In fact, in a IALM the d-set (and accordingly the state space and transition functions) evolves over time.
	\end{proof}
\end{customthm}
\begin{customlemma}{1}
	Consider the IALMs transitions $\mathcal{T}$ and $\hat {\mathcal{T}}$. For any $t$, augmented state $\bar s^t=(x^t,d^{t})=((x_{\text{loc}}^t,x_{\text{dest}}^t),d^{t})$ and action $a^{t}$,
	\begin{align*}
	\begin{split}
	&||(\mathcal{T}(\cdot\mid\bar s^t,a^t)- \hat {\mathcal{T}}(\cdot\mid\bar s^t,a^t)||_{1}\leq ||(I^t(\cdot\mid d^t)- \hat {I}^t(\cdot\mid d^t)||_{1}.
	\end{split}
	\end{align*}
	\begin{proof}
		According to the expression for the IALM transitions \eqref{IALMTransitions}
		\begin{align*}
		|\mathcal{T}(\bar s^{t+1}|\bar s^{t}, a^{t})-\hat{\mathcal{T}}(\bar s^{t+1}|\bar s^{t}, a^{t})|&= T(x_{\text{loc}}^{t+1}|x^{t},a^t)\,\left |\sum_{y_{\textit{src}}^t}T(x_{\text{dest}}^{t+1}|x^{t},y_{\textit{src}}^t,a^t)\,[I^t(y_{\textit{src}}^t|d^t)-\hat I^t(y_{\textit{src}}^t|d^t)]\right |\\
		& \leq T(x_{\text{loc}}^{t+1}|x^{t},a^t)\,\sum_{y_{\textit{src}}^t}T(x_{\text{dest}}^{t+1}|x^{t},y_{\textit{src}}^t,a^t)\,|I^t(y_{\textit{src}}^t|d^t)-\hat I^t(y_{\textit{src}}^t|d^t) |.
		\end{align*}
		Therefore,
		\begin{align*}
		||(\mathcal{T}(\cdot|\bar s^t,a^t)- \hat {\mathcal{T}}(\cdot|\bar s^t,a^t)||_{1}&=\sum_{\bar s^{t+1}}|\mathcal{T}(\bar s^{t+1}|\bar s^{t}, a^{t})-\hat{\mathcal{T}}(\bar s^{t+1}|\bar s^{t}, a^{t})|
		\\
		&\leq \sum_{\substack{x_{\text{loc}}^{t+1},x_{\text{dest}}^{t+1}}}T(x_{\text{loc}}^{t+1}|x^{t},a^t)\,\sum_{y_{\textit{src}}^t}T(x_{\text{dest}}^{t+1}|x^t,y_{\textit{src}}^t,a^t)\,|I^t(y_{\textit{src}}^t|d^t)-\hat I^t(y_{\textit{src}}^t|d^t) |\\
		&= \sum_{y^t_{\text{src}}}|I^t(y^t_{\text{src}}|d^t)-\hat I^t(y^t_{\text{src}}|d^t)|\sum_{x_{\text{loc}}^{t+1}}T(x_{\text{loc}}^{t+1}|x^{t},a^t)\sum_{x_{\text{dest}}^{t+1}}T(x_{\text{dest}}^{t+1}|x^{t},y^t_{\text{src}},a^t)\,\\
		&=\sum_{y^t_{\text{src}}}|I^t(y^t_{\text{src}}|d^t)-\hat I^t(y^t_{\text{src}}|d^t)|\\
		&=||I^t(\cdot\mid d^t)- \hat {I}(\cdot\mid d^t)||_{1}.
		\qedhere
		\end{align*}
	\end{proof}
\end{customlemma}
\begin{customthm}{2}
	Consider an IALM $\mathcal{M}=(\mathcal{S},A,\mathcal{T},R,h,b^0)$ and an AIP $\hat I$ inducing $\hat{\mathcal{M}}=(\mathcal{S},A,\hat{\mathcal{T}},R,h,b^0)$.
	Then, a value loss bound in terms of the $1$-norm error is given by
	\begin{equation*}
	||\mathcal{V}^*-\mathcal{V}^{\hat\pi^*}||_{\infty}\leq 2h^2 |R| \max_{t,d^{t}} ||(I^t(\cdot|d^t)- \hat {I}(\cdot|d^t)||_{1}.
	\end{equation*}
	\begin{proof}
		It suffices to apply Theorem \ref{ThTransitions} and Theorem \ref{TransitionsVSInfluence} consecutively to derive
		\begin{align*}
		||\mathcal{V}^*-\mathcal{V}^{\hat\pi^*}||_{\infty}&\leq 2h^2|R|\, \max_{t}\max_{\bar s^t,a^t}||(\mathcal{T}(\cdot\mid\bar s^t,a^t)- \hat {\mathcal{T}}(\cdot\mid\bar s^t,a^t)||_{1}
		\leq 2h^2 |R| \max_{t}\max_{d^t}||(I^t(\cdot\mid d^t)- \hat {I}^t(\cdot\mid d^t)||_{1}. \qedhere
		\end{align*}
	\end{proof}
\end{customthm}
\begin{customcor}{1}
	Consider a IALM $\mathcal{M}=(\mathcal{S},A,\mathcal{T},R,h,b^0)$ and an AIP $\hat I$ inducing $\hat{\mathcal{M}}=(\mathcal{S},A,\hat{\mathcal{T}},R,h,b^0)$.
	Then, a value loss bound in terms of KL divergence error is given by
	\begin{equation*}
	||\mathcal{V}^*-\mathcal{V}^{\hat\pi^*}||_{\infty}\leq 2h^2 |R| \max_{t,d^{t}} \sqrt{2 D_{KL}(I^t(\cdot|d^t)||\hat I^t(\cdot|d^t))}.
	\end{equation*}
	\begin{proof}
		It suffices to apply Pinsker inequality \cite{cover2012elements} to equation \eqref{InfluenceBoundEq} to obtain the claim
		\begin{align*}
		||\mathcal{V}^*-\mathcal{V}^{\hat\pi^*}||_{\infty}&\leq 2h^2 |R| \max_{t,d^{t}} ||(I^t(\cdot|d^t)- \hat {I}(\cdot|d^t)||_{1}
		\leq 2h^2 |R| \max_{t,d^{t}} \sqrt{2D_{KL}(I^t(\cdot|d^t)||\hat I^t(\cdot|d^t))}.
		\qedhere
		\end{align*}
	\end{proof}
\end{customcor}
We now introduce a Lemma
from \cite{weissman2003inequalities} that allows to bound the $L^1$ distance between the empirical and the actual distribution in probability in terms of the sample size $N$.  
\begin{lemma} \cite{weissman2003inequalities}\label{L1Prob}
	Let $p$ be a probability distribution over $S$ and $p_N$ the empirical distribution drawn from a sample of size $N$. For every $\varepsilon>0$
	$$
	\mathbb{P}(||p-p_N||_1\geq \varepsilon)\leq (2^{|S|}-2)e^{-N\frac{\varepsilon^2}{2}}.
	$$
\end{lemma}
We will leverage this result for the proof of Theorem \ref{ProbBound}.
\begin{customthm}{3}
	Consider a IALM $\mathcal{M}=(\mathcal{S},A,\mathcal{T},R,h,b^0)$ and an AIP $\hat I$ inducing $\hat{\mathcal{M}}=(\mathcal{S},A,\hat{\mathcal{T}},R,h,b^0)$. Assume that for every time $t$ and d-set instantiation $d^t$, we have at least an influence sources sample of size $N$.
	Then  for every $\varepsilon >0$
	\begin{align*}
	\mathbb{P}\left (||\mathcal{V}^*-\mathcal{V}^{\hat\pi^*}||_\infty\leq 2 h^2 |R|\max_{t,d^t}||(I^t_N(\cdot|d^t)- \hat {I}^t(\cdot|d^t)||_{1}+\varepsilon\right )\geq 1-h\,|D^h|(2^{|Y_{\text{src}}|}-2)e^{-N\frac{\varepsilon^2}{2}},
	\end{align*}
	where $|Y_{\text{src}}|$ refers to the cardinality of the influence sources space and $|D^h|$ the cardinality of the $h$ step d-sets space.
	\begin{proof}
		We apply Lemma \ref{L1Prob} from \cite{weissman2003inequalities} to the exact influence $I^t$ and the empirical influence $I^t_N$, thereby obtaining 
		\begin{align*}
		\mathbb{P}&\left (||I^t(\cdot\mid d^t)- I^t_N(\cdot\mid d^t)||_{1}\geq \varepsilon\right )
		\leq (2^{|Y_{\text{src}}|}-2)e^{N\frac{\varepsilon^2}{2}}.
		\end{align*}
		Using the above inequality and the reverse triangle inequality
		\begin{align*}
		&\left |||I^t(\cdot\mid d^t)- \hat I^t(\cdot\mid d^t)||_{1}-||\hat I^t(\cdot\mid d^t)- I^t_N(\cdot\mid d^t)||_{1}\right |\leq ||I^t(\cdot\mid d^t)- I^t_N(\cdot\mid d^t)||_{1},
		\end{align*}
		we obtain
		\begin{align*}
		\mathbb{P}\left (||I^t(\cdot\mid d^t)- \hat I^t(\cdot\mid d^t)||_{1}\geq ||\hat I^t(\cdot\mid d^t)- I^t_N(\cdot\mid d^t)||_{1}+ \varepsilon\right )&\leq
		\mathbb{P}\left (\,\left |||I^t(\cdot\mid d^t)- \hat I^t(\cdot\mid d^t)||_{1}-||\hat I^t(\cdot\mid d^t)- I^t_N(\cdot\mid d^t)||_{1}\right |\geq \varepsilon\right )\\
		&\leq 
		\mathbb{P}\left (\,||I^t(\cdot\mid d^t)- I^t_N(\cdot\mid d^t)||_{1})\geq \varepsilon\right )\leq (2^{|Y_{\text{src}}|}-2)e^{N\frac{\varepsilon^2}{2}}.
		\end{align*}
		Thus by the union bound,
		\begin{align}
		\label{3}
		\mathbb{P}&\left (\max_{t,d^t} ||I^t(\cdot\mid d^t)- \hat I^t(\cdot\mid d^t)||_{1}\leq \max_{t,d^t}||I^t_N(\cdot\mid d^t)- \hat I^t(\cdot\mid d^t)||_{1}+\varepsilon \right )\nonumber\\
		&\geq\nonumber\mathbb{P}\left (\bigcap_{t,d^t} \left \{ ||I^t(\cdot\mid d^t)- \hat I^t(\cdot\mid d^t)||_{1}\leq ||I^t_N(\cdot\mid d^t)- \hat I^t(\cdot\mid d^t)||_{1}+\varepsilon \right \}\right )\nonumber\\
		&=1-\mathbb{P}\left (\bigcup_{t,d^t} \left \{ ||I^t(\cdot\mid d^t)- \hat I^t(\cdot\mid d^t)||_{1}\geq ||I^t_N(\cdot\mid d^t)- \hat I^t(\cdot\mid d^t)||_{1}+\varepsilon \right \} \right )\nonumber\\
		& \geq 1-\sum_{t,d^t}\mathbb{P}\left ( ||I^t(\cdot\mid d^t)- \hat I^t(\cdot\mid d^t)||_{1}\geq ||I^t_N(\cdot\mid d^t)- \hat I^t(\cdot\mid d^t)||_{1}+\varepsilon \right )\nonumber\\
		& \geq 1- \sum_{t,d^t}(2^{|Y_{\text{src}}|}-2)e^{-N\frac{\varepsilon^2}{2}}\geq 1-h\,|D^h|\,(2^{|Y_{\text{src}}|}-2)e^{-N\frac{\varepsilon^2}{2}}.
		\end{align}
		Now combining Theorem \ref{ValueVSInfluence} and \eqref{3}, we get the statement.
	\end{proof}
\end{customthm}
\section{Experimental setup}
Here we describe the experimental setup for the domains we consider in Section \ref{experiments}.  
\subsection{Planetary Exploratory}
The setting is a version of Planetary Exploration \cite{witwicki2010influence}, in which a planetary rover has to explore an area. At the same time a second agent, a satellite, may help the rover by planning a route, which gives a higher probability of a successful move of the rover. Figure \ref{roverfig} in the paper shows a depiction of the three time steps dynamic Bayesian network \cite{murphy2002dynamic}. If the rover fails to move it receives a penalty of $-0.5$ and reaching the target position results in a reward of $10$. The satellite action of helping or not helping is denoted by $a_\text{sat} \in \{0,1\}$, while the (non)-availability of a routing plan is a binary factor $pl \in \{\text{plan}, \neg \text{plan}\}$. The action $a_{\text{sat}}$ of the satellite depends on its own charge status, a resource which it has to manage as well. For the results presented in Figure \ref{MR}, the satellite employs a stochastic policy such that whenever the level of the battery is lower than a given threshold, it will try to provide a plan with some probability to fail. 
In this domain we set the horizon to $6$ and we will keep the initial state distribution fixed.

The rover does not observe the satellite charge, but it can use the history of $pl$ as d-set to predict the satellite actions $a_\text{sat}$ which constitutes the influence sources. Thus, we try to predict the influence $I^t(a^t_\text{sat} \mid (pl^0,\cdots,pl^t))$ at time $t$ based on the history of $pl$ up to time $t$.

For this approximation we train a LSTM neural network (NN) \cite{LSTM} up to 15 epochs with 10000 samples drawn from a exploratory random policy for the rover. 
At each train time the NN sees the full history of plans $(pl^0,\cdots,pl^h)$
and predicts an approximate influence point
$ \{ \hat{I}^t ( \cdot \mid pl^0,\cdots,pl^t ) \}_{t=0:h} $.

\subsection{Traffic Network}
We simulate a busy traffic network represented by a regular grid formed by $4$ road segments and $4$ traffic light agents at the intersections (see Figure \ref{TrafficGrid}).
At every time step, each agent observes the state of a $3\times 3$ grid around the intersection. We consider the traffic light at position $(1,1)$ as the local agent. We employ fixed policies for the other $3$ agents. Precisely, agent $(1,4)$ and agent $(4,1)$ prioritize the horizontal lane whenever a vehicle is waiting. For the agent in $(4,4)$ we use a random policy. 
The local space of the agent is composed by the observations at each side of the intersection $X=(south,\, east,\, north,\, west)$, for $south,\, east,\, north,\, west\in\{0,1\}$ where $1$ corresponds to a vehicle and $0$ to none. The local agent can then decide to prioritize the horizontal or the vertical lane $a_{loc}\in\{hor,vert\}$. Then, it will receive a local reward corresponding to the number of cars waiting at the intersection $R(x)=-east-north$. The outgoing vehicles from $west$ and $south$ have some probability to re-enter in the system from the parallel lane and then affect the decisions at other intersections.
In this domain we set the horizon to $4$ and we will keep the initial state distribution fixed.
The only information the local agent needs to retrieve from the rest of the model is whether a car will access the local model from the incoming lanes. That is $X_{\text{dest}}=(east,north)$. To predict the car inflow, the local necessary information consists of the history of outgoing cars from $south$ and $west$, $D^t=((south^0,\, east^0),\dots,(south^t,\, east^t))$.

For this approximation we train a LSTM NN up to 16 epochs with 10000 samples drawn from a exploratory random policy for the local agent. At each training time the NN sees the history of the $(south,\, east)$ and predicts the distribution of $(east^{t},north^{t})$.
\subsection{Fire Fighters Problem}
In the Fire Fighters problem \cite{oliehoek2010value}, we model a team of 2 agents that have to extinguish fires in a row of 3 houses. At every time step, each agent can choose to fight fires at one of its neighboring houses (see Figure \ref{FFP}). The actions of the two agents are denoted by $a_1,a_2\in \{\text{right},\text{left}\}$.
\begin{figure}[t]
	\begin{subfigure}{0.45\textwidth}
		\includegraphics[scale=0.55]{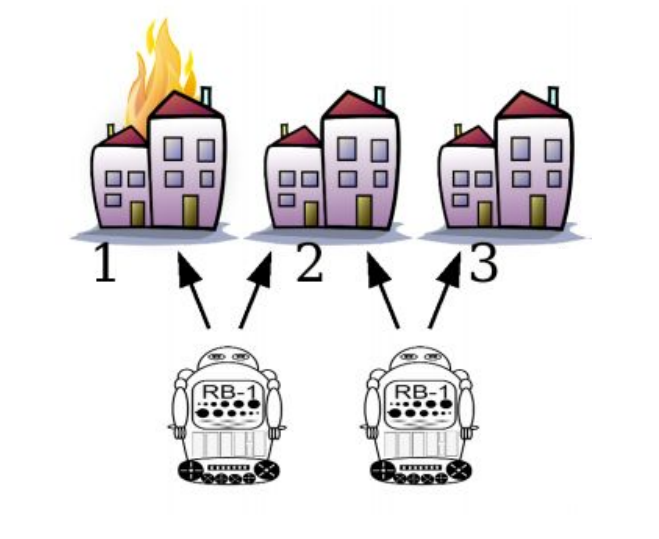}
		\caption{An illustration of the fire fighters problem.}
		\label{FFP}
	\end{subfigure}
	\begin{subfigure}{0.45\textwidth}
		\includegraphics[scale=0.5]{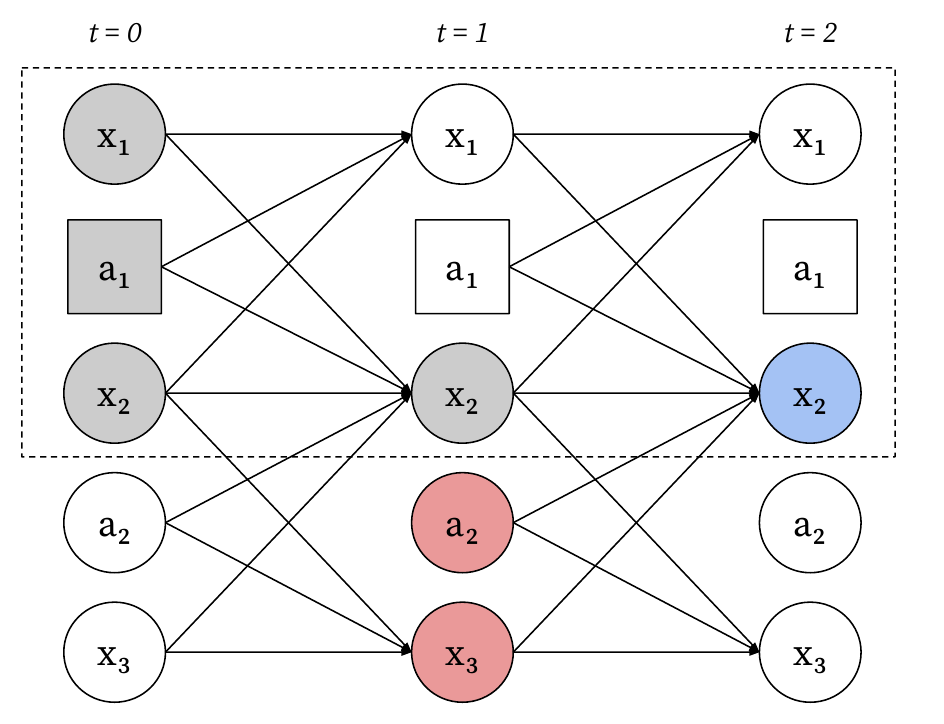}
		\caption{A dynamic Bayesian network of the fire fighters domain. The dotted square refers to the local model of agent $1$. The gray circles constitute the d-set $D^1$ for the influence sources $a^1_2,x^1_3$ at time $t=1$ depicted as red circles. The blue variable $x^2_2$ represents the influence destination, i.e. the modeled factor that experiences directly the external influence.}
		\label{2DBN_FFP}
	\end{subfigure}
	\caption{
		Fire Fighters.}
\end{figure}
The state space consists of three binary factors $x_1,x_2,x_3\in\{0,1\}$, representing the fire level of the houses, where $1$ indicates a burning house. Thus, a state corresponds to an assignment of the tuple $s=(x_1,x_2,x_3)$.
If a house is not burning and no fire fighting agent is present with probability $0.9$ the fire can spread from a neighboring house. A single agent present at a house will succeed in extinguishing the fires with probability $1$ if none of the neighboring houses are burning and with probability $0.6$ otherwise. If two agents fight fires at the same house, they extinguish the fire with probability $1$. 

We consider the local model for agent $1$ including only houses $x_1,x_2$. In fact, it can only observe its neighboring houses $x_1,x_2$ and receives a reward according to the number of local houses that are burning at the next time step $R(s')=-x'_1-x'_2$. Figure \ref{2DBN_FFP} shows an illustration of the dynamic Bayesian network.
In this domain we set the horizon to $4$ and we will keep the initial state distribution fixed.

The local agent does not observe the state of house $x_3$, therefore it can not predict the action of the non-local agent $a_2$. However, the history of factors $x_1, x_2$, action $a_1$ constitute the d-set for the influence sources $a_2, x_3$. Thus we try to predict the influence sources $a^t_2, x^t_3$ given the d-set $D^t=(x^0_1,a^0_1,x^0_2,\dots,x^{t-1}_1,a^{t-1}_1,x^{t-1}_2,x^t_2)$ at time $t$. 

For this approximation we train a LSTM NN up to 20 epochs with 100000 samples drawn from a exploratory random policy for the local agent $1$. 
At each training time the NN sees the history of the houses $x_1,x_2$ and the actions taken by the local agent $a_1$ and predicts an approximate influence point
$ \{ \hat{I}^t ( \cdot \mid (x^0_1,a^0_1,x^0_2,\dots,x^{t-1}_1,a^{t-1}_1,x^{t-1}_2,x^t_2) ) \}_{t=0:h} $.
\section{Additional results}
Figure \ref{MR_random} and Figure \ref{MR_deterministic} shows further results on the Planetary rover exploration domain. Precisely they present the value, test errors and correlations for the rover best response against two different policies of the satellite agent. In Figure \ref{MR_random}, the satellite is randomly deciding at any time step to provide or not a plan. In Figure \ref{MR_deterministic} the satellite decides deterministically to provide a plan whenever the level of the battery is lower than a given threshold. In both cases, the performances of the approximate-influence policy $\hat\pi^*$ improve with the training epochs and the cross entropy error is highly correlated with the value loss.
\begin{figure*}[!h]
	\begin{subfigure}{0.25\textwidth}
		\includegraphics[scale=0.3]{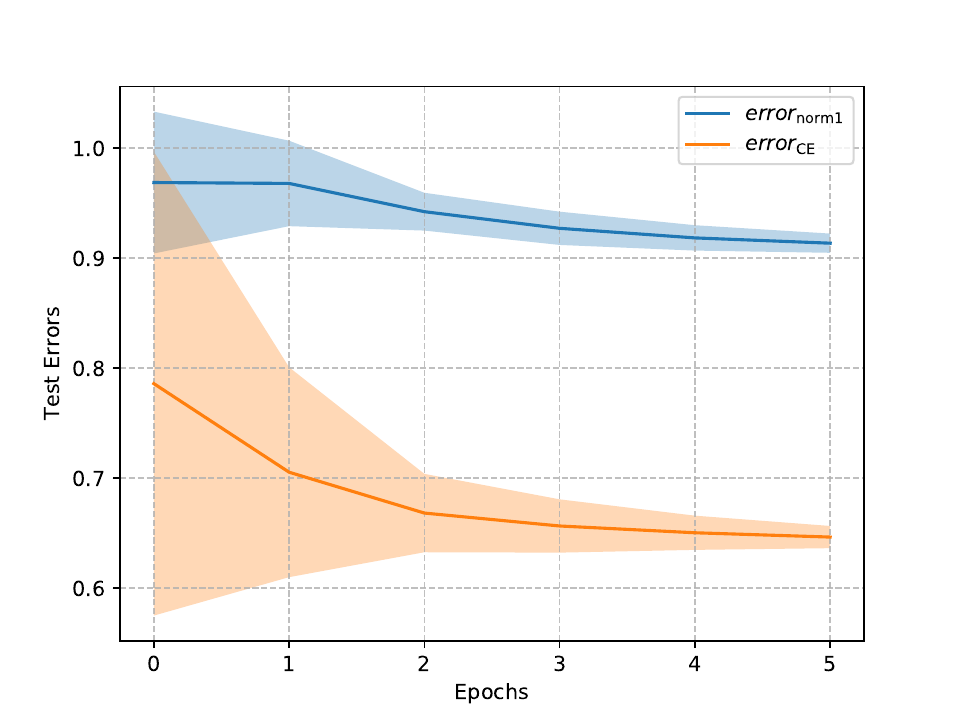}
	\end{subfigure}
	\begin{subfigure}{0.25\textwidth}
		\includegraphics[scale=0.3]{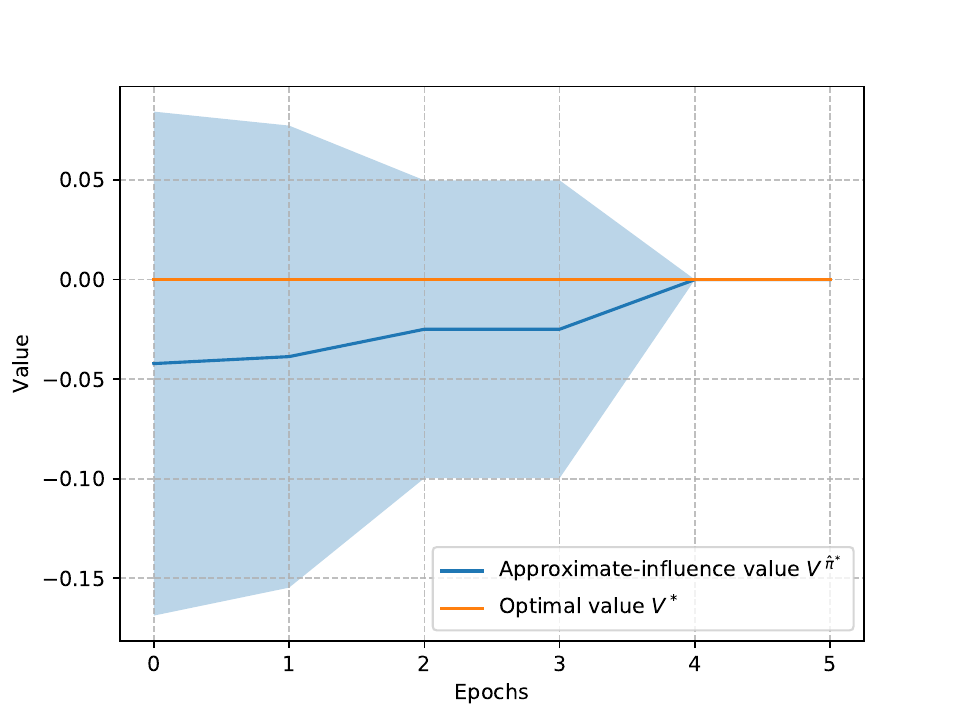}
	\end{subfigure}
	\begin{subfigure}{0.25\textwidth}
		\includegraphics[scale=0.3]{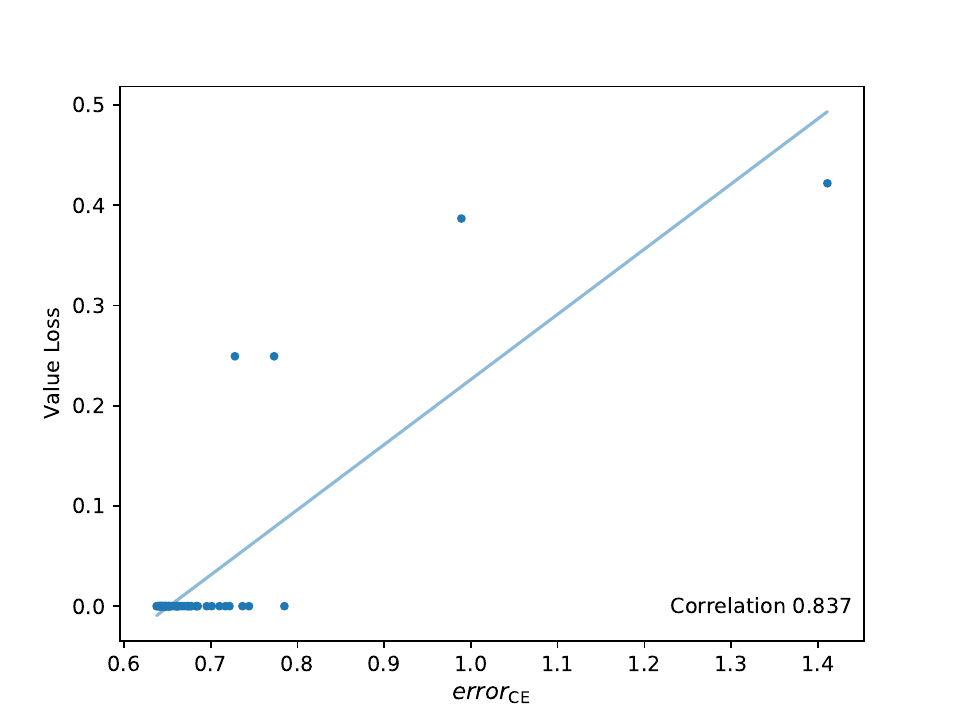}
	\end{subfigure}
	\begin{subfigure}{0.23\textwidth}
		\includegraphics[scale=0.3]{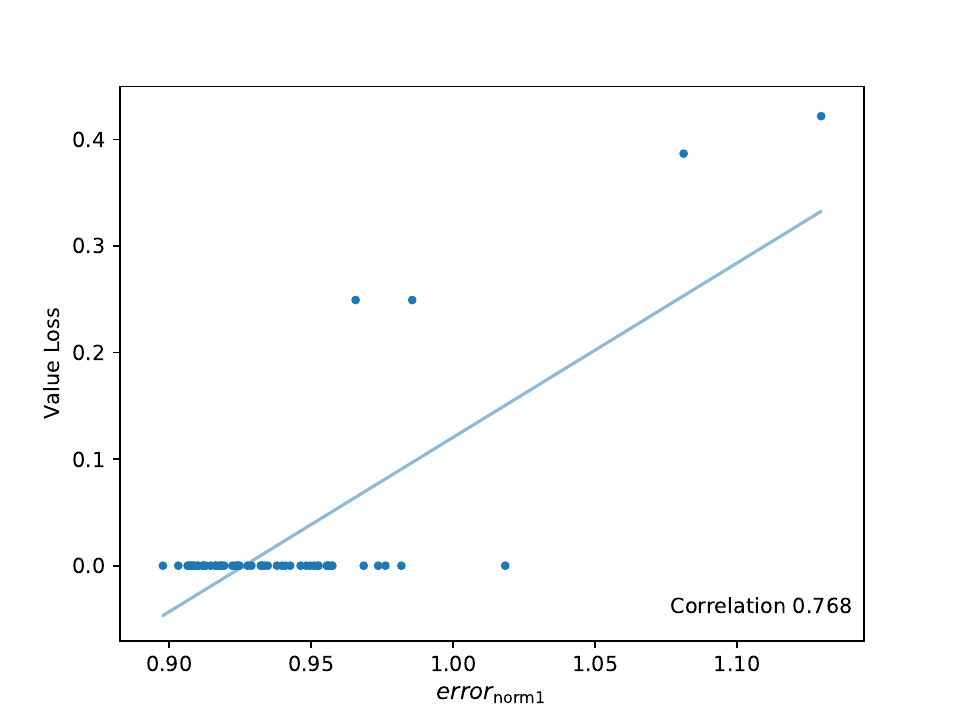}
	\end{subfigure}
	\caption{Random policy of the satellite} 
	\label{MR_random}
\end{figure*}
\begin{figure*}[!h]
	\begin{subfigure}{0.25\textwidth}
		\includegraphics[scale=0.3]{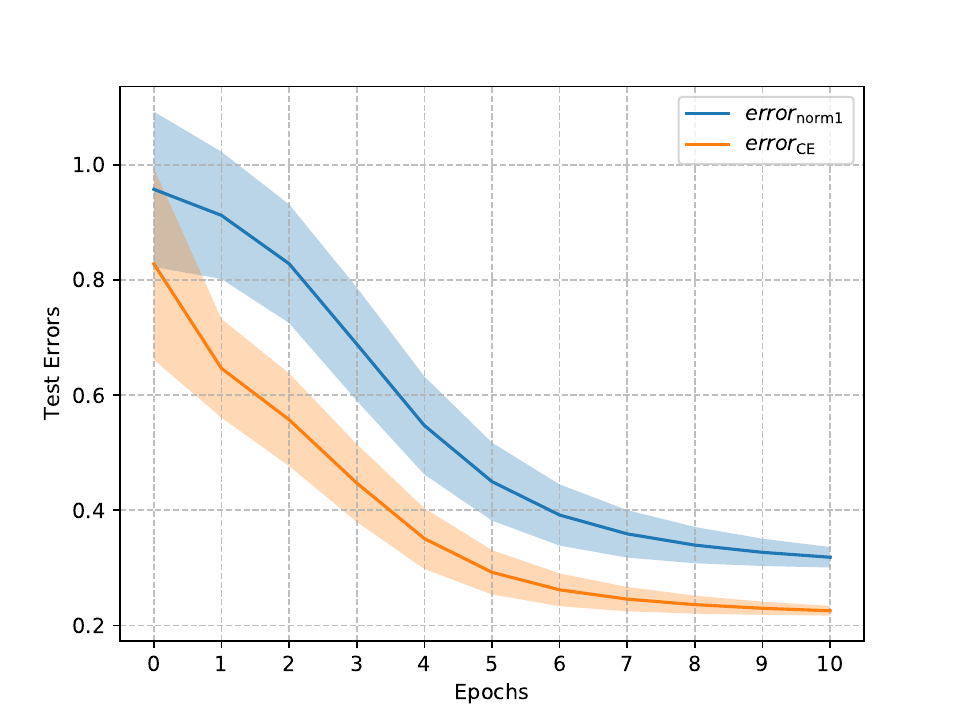}
	\end{subfigure}
	\begin{subfigure}{0.25\textwidth}
		\includegraphics[scale=0.3]{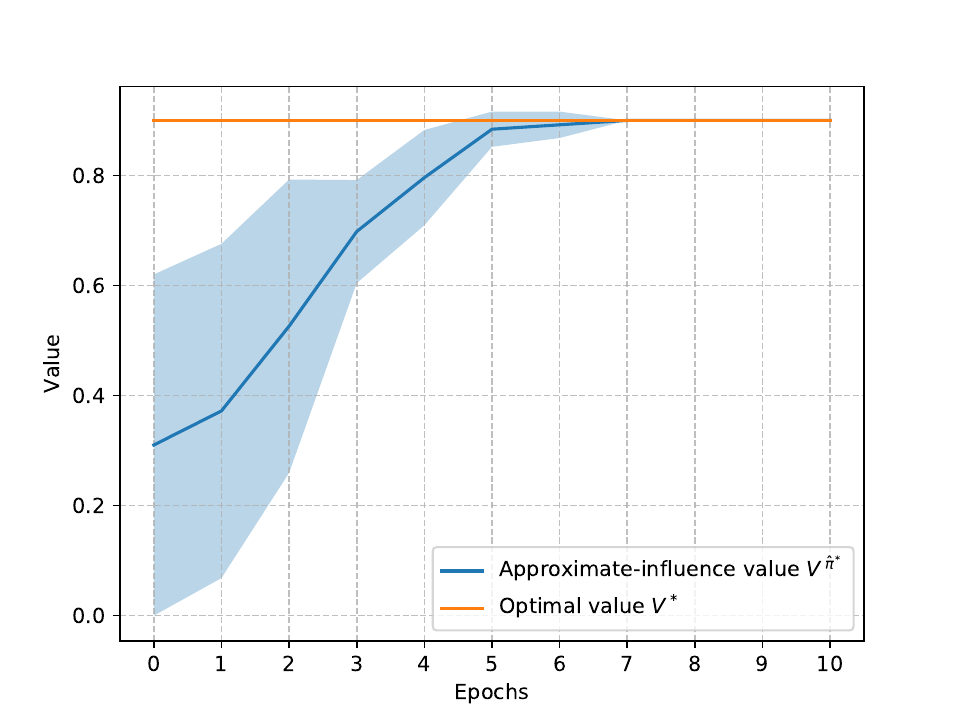}
	\end{subfigure}
	\begin{subfigure}{0.25\textwidth}
		\includegraphics[scale=0.3]{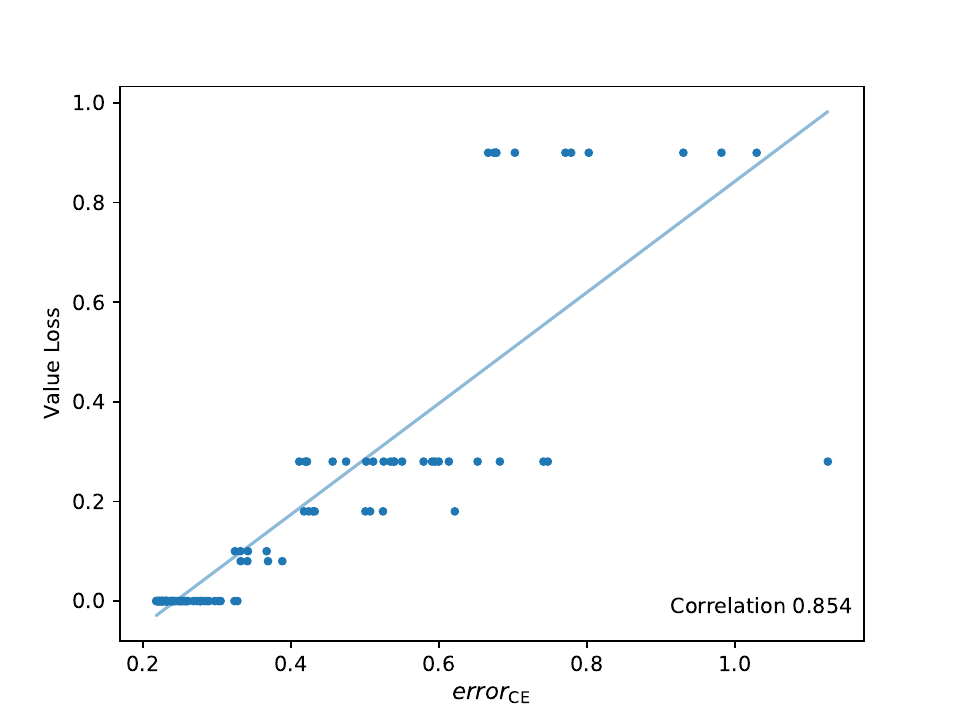}
	\end{subfigure}
	\begin{subfigure}{0.23\textwidth}
		\includegraphics[scale=0.3]{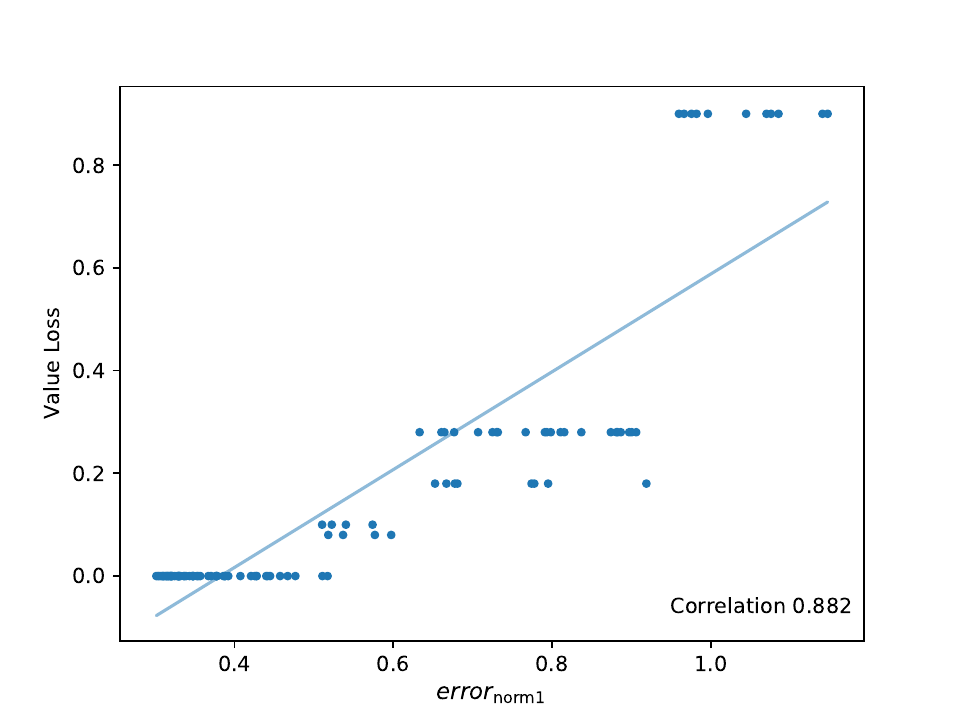}
	\end{subfigure}
	\caption{Deterministic policy of the satellite.} 
	\label{MR_deterministic}
\end{figure*}
\end{flushleft}
\end{document}